\documentclass[sn-mathphys-num,iicol]{sn-jnl}

\usepackage{graphicx}%
\usepackage{multirow}%
\usepackage{amsmath,amssymb,amsfonts}%
\usepackage{amsthm}%
\usepackage{mathrsfs}%
\usepackage[title]{appendix}%
\usepackage{textcomp}%
\usepackage{manyfoot}%
\usepackage{booktabs}%
\usepackage[ruled,vlined]{algorithm2e}
\usepackage{listings}%

\usepackage{setspace}
\usepackage{bm}
\usepackage{lineno}
\usepackage{cases}
\usepackage{psfragx}
\usepackage{psfrag}
\usepackage{pifont}
\usepackage[table]{xcolor}
\usepackage{makecell}
\usepackage{url}
\usepackage{stmaryrd}
\usepackage{arydshln}
\usepackage{mathrsfs}
\usepackage{verbatim}
\usepackage{enumerate}
\usepackage{lipsum}
\usepackage{booktabs}
\usepackage{threeparttable}
\usepackage[normalem]{ulem}
\usepackage{mathtools}
\usepackage{wrapfig}
\usepackage{tablefootnote}
\usepackage{diagbox}
\usepackage{bbding}
\usepackage{subcaption}
\usepackage{hvfloat}

\theoremstyle{thmstyleone}%
\theoremstyle{thmstyletwo}%

\theoremstyle{thmstylethree}%

\newcommand{\eg}{e.g.}
\newcommand{\ie}{i.e.}
\newcommand{\wrt}{w.r.t.}

\newcommand{\etc}{etc.}

\SetCommentSty{mycommfont}
\definecolor{bleudefrance}{rgb}{0.19, 0.55, 0.91}
\definecolor{caribbeangreen}{rgb}{0.0, 0.8, 0.6}

\usepackage{amsmath,amsfonts,bm}

\def\eqref#1{equation~\ref{#1}}

\def\1{\bm{1}}

\def\vs{{\bm{s}}}

\def\mA{{\bm{A}}}

\def\mD{{\bm{D}}}
\def\mE{{\bm{E}}}

\def\mG{{\bm{G}}}

\DeclareMathAlphabet{\mathsfit}{\encodingdefault}{\sfdefault}{m}{sl}
\SetMathAlphabet{\mathsfit}{bold}{\encodingdefault}{\sfdefault}{bx}{n}

\newcommand{\softmax}{\mathrm{softmax}}

\input{abbr_commands.tex}

\begin{document}

\title{An Experimental Study on Exploring Strong Lightweight Vision Transformers via Masked Image Modeling Pre-Training}


\author[1,2]{\fnm{Jin} \sur{Gao}}\email{jin.gao@nlpr.ia.ac.cn}
\equalcont{These authors contributed equally to this work.}

\author[1,2]{\fnm{Shubo} \sur{Lin}}\email{linshubo2023@ia.ac.cn}
\equalcont{These authors contributed equally to this work.}

\author*[1]{\fnm{Shaoru} \sur{Wang}}\email{wangshaoru2018@ia.ac.cn}
\equalcont{These authors contributed equally to this work.}

\author[1,2]{\fnm{Yutong} \sur{Kou}}\email{kouyutong2021@ia.ac.cn}

\author[4]{\fnm{Zeming} \sur{Li}}\email{lizeming@megvii.com}

\author[5]{\fnm{Liang} \sur{Li}}\email{liang.li.brain@aliyun.com}

\author[6]{\fnm{Congxuan} \sur{Zhang}}\email{zcxdsg@163.com}

\author[7]{\fnm{Xiaoqin} \sur{Zhang}}\email{zhangxiaoqinnan@gmail.com}

\author[5]{\fnm{Yizheng} \sur{Wang}}\email{yzwang57@sina.com}

\author[1,2,3]{\fnm{Weiming} \sur{Hu}}\email{wmhu@nlpr.ia.ac.cn}

\affil[1]{\orgdiv{State Key Laboratory of Multimodal Artificial Intelligence System}, \orgname{Institute of Automation, Chinese Academy of Sciences}, \orgaddress{\postcode{100190}, \state{Beijing}, \country{China}}}

\affil[2]{\orgdiv{School of Artificial Intelligence}, \orgname{University of Chinese Academy of Sciences}, \orgaddress{\postcode{101408}, \state{Beijing}, \country{China}}}

\affil[3]{\orgdiv{School of Information Science and Technology}, \orgname{ShanghaiTech University}, \orgaddress{\postcode{201210}, \state{Shanghai}, \country{China}}}

\affil[4]{\orgname{Megvii Research}, \orgaddress{\postcode{100089}, \state{Beijing}, \country{China}}}

\affil[5]{\orgname{Beijing Institute of Basic Medical Sciences}, \postcode{100850}, \state{Beijing}, \country{China}}

\affil[6]{\orgname{Nanchang Hangkong University}, \postcode{330063}, \state{Beijing}, \country{China}}

\affil[7]{\orgname{Zhejiang University of Technology}, \postcode{310014}, \state{Hangzhou}, \country{China}}



\abstract{Masked image modeling (MIM) pre-training for large-scale vision transformers (ViTs) has enabled promising downstream performance on top of the learned self-supervised ViT features. In this paper, we question if the \textit{extremely simple} lightweight ViTs' fine-tuning performance can also benefit from this pre-training paradigm, which is considerably less studied yet in contrast to the well-established lightweight architecture design methodology. We use an observation-analysis-solution flow for our study. We first systematically \textbf{observe} different behaviors among the evaluated pre-training methods with respect to the downstream fine-tuning data scales. Furthermore, we \textbf{analyze} the layer representation similarities and attention maps across the obtained models, which clearly show the inferior learning of MIM pre-training on higher layers, leading to unsatisfactory transfer performance on data-insufficient downstream tasks. This finding is naturally a guide to designing our distillation strategies during pre-training to \textbf{solve} the above deterioration problem. Extensive experiments have demonstrated the effectiveness of our approach. Our pre-training with distillation on pure lightweight ViTs with vanilla/hierarchical design ($5.7M$/$6.5M$) can achieve $79.4\%$/$78.9\%$ top-1 accuracy on ImageNet-1K. It also enables SOTA performance on the ADE20K segmentation task ($42.8\%$ mIoU) and LaSOT tracking task ($66.1\%$ AUC) in the lightweight regime. The latter even surpasses all the current SOTA lightweight CPU-realtime trackers.}

\keywords{Masked image modeling, vision transformers, lightweight networks, knowledge distillation, visual tracking}



\maketitle

\section{Introduction}\label{sec:introduction}

{Self-supervised} learning (SSL) has shown great progress in representation learning without heavy reliance on expensive labeled data. SSL focuses on various pretext tasks for pre-training, the objectives of which provide a richer learning signal (\eg, richer visual information in computer vision field) than the supervised objective of predicting a single label or concept selected from a predefined set of different categories~\cite{dino}. This can explain why several existing SSL works~\cite{mocov1, mocov2, BYOL, swav, mocov3, dino} based on contrastive learning (CL) have managed to achieve comparable or even better accuracy than fully-supervised pre-training when transferring the learned representations based on ImageNet-1K~\cite{ImageNet} to downstream tasks. Prior works~\cite{SEED, CompRess, oss} have also sought to apply CL on lightweight convolutional networks (ConvNets) and improve the performance by distillation, because efficient networks are essential for modern on-device computer vision.

Recently, another SSL trend focuses on masked image modeling (MIM) pre-training (\eg,~\cite{beit, mae, ibot, hiera}), which perfectly fits vision transformers (ViTs)~\cite{vit} for vision tasks and has shown emerging good generalization performance and scalability to large-scale models and datasets. Although supervised pre-training for ViTs can also benefit from the ``LLM-like" scaling in vision, \ie, scaling up training duration, model architecture size and number of training images together to improve representation quality, it has also been demonstrated that lightweight ViTs can hardly benefit from either the largest dataset or compute resources for supervised training due to the reason that their representation quality is bottlenecked by the limited capacity. In other words, the performance improvement of scaling up compute and data for supervised training of lightweight ViTs may be marginal considering the substantially increased cost of computation or collecting and labeling the data. For instance, scaling up the supervised pre-training duration on ImageNet-21K~\cite{ImageNet} (a bigger and more diverse dataset, as roughly ten times the size of ImageNet-1K) from 20 to 200 hours even for the lightweight specialized Data-efficient image Transformers (dubbed DeiT)~\cite{deit} can only achieve $+0.9\%$ top-1 accuracy gain on ImageNet in our experiment. We also empirically examine whether the self-supervised pre-training on lightweight ViTs can benefit from large-scale pre-training data. A similar phenomenon to the supervised pre-training is observed that the benefit of using large-scale data is limited due to the limited capacity of lightweight ViTs. It is thus appealing to explore other factors in improving the representation quality of lightweight ViT-based models, \eg, exploring the design space for ViT-based architectures or shifting from ``LLM-like" scaling of data to SSL pre-training with optimized pretext objectives. 

\begin{figure}[!t]
    \centering
    \subfloat{\includegraphics[width=1\linewidth]{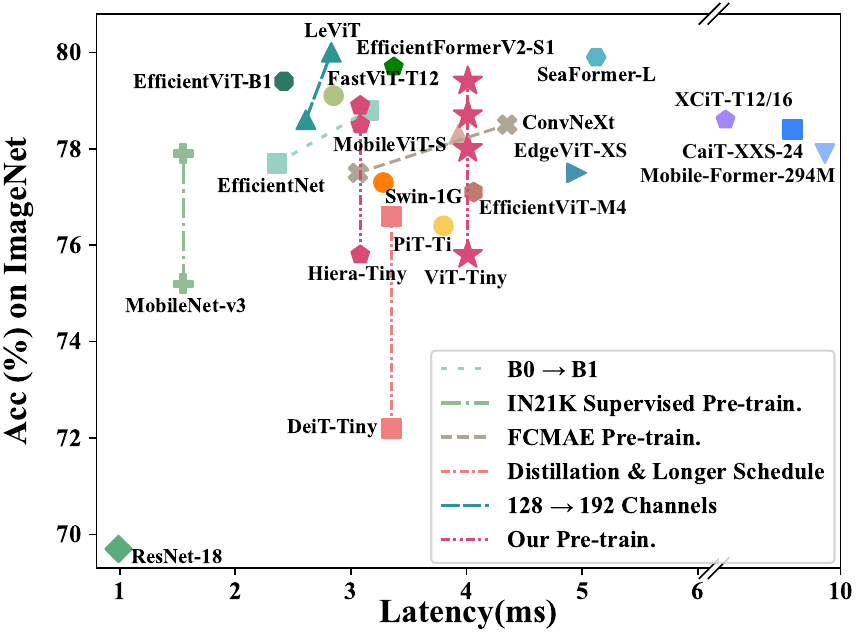}
    \label{SOTAcomparison}}
    \vspace{0pt}
    \caption{Our SSL pre-training with distillation on pure lightweight ViT-Tiny (5.7M)/Hiera-Tiny (6.5M) can achieve $79.4\%$/$78.9\%$ top-1 accuracy on ImageNet-1K validation set, which bridges the performance gap between \emph{extremely simple} ViT architectures and delicately designed ones in the lightweight regime. The latency is measured on Orin with batch size 1. The transfer evaluation on other downstream tasks is also impressive.} \label{fig:SOTAcomparison}
    \vspace{-5pt}
    \end{figure}

A significant body of work has been devoted to designing different lightweight ViT derivatives by introducing sophisticated components with inductive biases~\cite{levit, xcit, pit, cait, mobilevit, mobileformer, edgevit, fastvit, efficientformerv2, efficientvit-iccv, seaformer}, leading to top-performing lightweight models, while little attention is paid to how to optimize the pre-training strategies for further improving the existing lightweight models, especially the \emph{extremely simple} ones with plain architecture or less artificial design. We believe the latter is also of vital importance and the utilization of MIM pre-training is one of the most promising approaches which echo this direction, since MIM has achieved great progress on large-scale models. To this end, we first develop and carefully adapt various recently popular MIM pre-training methods (mainly for the large models) to the above lightweight regime, \ie, the MAE~\cite{mae} and SimMIM~\cite{simmim} methods which aim to predict raw RGB pixels of masked patches, and some methods with other pre-defined targets for prediction (\eg, discrete tokens in BEiT~\cite{beit} which are generated by a dVAE~\cite{dvae} pre-trained on DALLE~\cite{dalle}, features from a momentum teacher in BootMAE~\cite{bootmae}, and HoG features in MaskFeat~\cite{maskfeat}). We then benchmark them along with two typical CL pre-training methods (\ie, MoCo-v3~\cite{mocov3} and DINO~\cite{dino}) and the fully-supervised pre-training methods as the baselines on ImageNet, as well as on some data-insufficient downstream classification tasks and dense prediction tasks (\eg, object detection and segmentation, semantic segmentation, and visual tracking), for a closer look at the self-supervised pre-training of lightweight ViTs. 

On one hand, we surprisingly find that \emph{if proper pre-training is adopted, even the extremely simple lightweight ViTs with vanilla design show comparable performance to the current SOTA ViT derivatives with delicate design on ImageNet}. Without bells and whistles, our implementation with MIM pre-training can achieve $78.0\%$ top-1 accuracy (see~\cref{tab:imagenetcompare}) by using a slightly modified version of~\cite{deit} with vanilla lightweight architecture, namely ViT-Tiny (5.7M) in this paper. This finding is intriguing since the results indicate that proper pre-training can bridge the performance gap between vanilla ViT architectures and delicately designed ones in the lightweight regime to a great extent.
Although the latter ones may also enjoy high accuracy and attractive floating point operation (FLOP) counts, some of them (\eg, EdgeViT-XS~\cite{edgevit}, MobileViT-S~\cite{mobilevit}, Mobile-Former-294M~\cite{mobileformer}, SeaFormer-L~\cite{seaformer}) with added complexity from specialized modules make the inference speed slower overall (higher latency, see~\cref{tab:sota}).
This interesting finding, however, does not match the practice of lightweight ViT pre-training in~\cite{tinymim}, which claims that MIM pre-training can even hurt lightweight ViT fine-tuning accuracy on ImageNet. On the other hand, we systematically observe different behaviors between MIM and CL pre-training with respect to the downstream fine-tuning data scales, \ie, MIM pre-training generally outperforms CL pre-training on the data-sufficient ImageNet classification task, while \emph{showing significantly worse performance on the data-insufficient downstream tasks}. 

These findings motivate us to delve into the working mechanism of the above self-supervised pre-training methods for lightweight ViTs. More specifically, we first analyze their frozen features under linear probing evaluation and then introduce a model analysis method to study the pattern of layer behaviors during pre-training and fine-tuning, \ie, analyzing the layer representation similarities across the obtained models, both of which clearly show the inferior learning of MIM pre-training on higher layers. In specific, MIM pre-training hardly learns semantics at an abstract level relevant to recognition in higher layers, which is contrary to the CL pre-training (\eg, DINO~\cite{dino} can automatically learn class-specific features leading to unsupervised object segmentation). This may be the reason for its unsatisfactory fine-tuning performance on data-insufficient downstream tasks. In other words, \emph{higher layers matter in data-insufficient downstream tasks}. This is also consistent with the further model analysis by studying the attention maps across the obtained models, which shows that MIM pre-training tends to make the pre-trained models focus on local patterns with concentrated attention in higher layers. By further carefully investigating what really matters for downstream performance, however, we find that \emph{lower layers of the pre-trained models matter more than higher ones if sufficient downstream data is provided}. We attribute it to the reason that \emph{MIM pre-training is prone to making the attention of the downstream models more local and concentrated} than the CL pre-training and training-from-scratch settings in some layers, \ie, drawing locality inductive bias from pre-training into the downstream models to make them focus on nearby image elements, which may be the key to the performance gain on the data-sufficient ImageNet classification task.

Based on the above analyses, we have developed a decoupled distillation strategy for better MIM pre-training of lightweight ViTs, which not only learns semantics at an abstract level relevant to recognition in higher layers (see~\cref{tbl:transfer-distill} for the significantly improved performance on data-insufficient downstream classification tasks and dense prediction tasks), but also makes the downstream models preserve the above useful locality inductive bias (see~\cref{tbl:transfer-distill} for the improved performance on the data-sufficient ImageNet classification task). We also experimentally demonstrate the effectiveness of our observation-analysis-solution flow by exploring a simple, efficient, lightweight, yet accurate hierarchical ViT from Hiera~\cite{hiera}, namely Hiera-Tiny (6.5M). In particular, our pre-training with distillation on the pure ViT-Tiny and Hiera-Tiny can achieve $79.4\%$/$78.9\%$ top-1 accuracy on ImageNet-1K, which is comparable to the current SOTA networks with sophisticated architecture design, including some high-performing lightweight ConvNets. It also improves the corresponding supervised pre-training baselines to achieve $+2.1$/$+5.3$ box AP and $+2.0$/$+4.4$ mask AP gains on the COCO~\cite{coco} detection and segmentation tasks. Our pre-training also enables SOTA performance on the ADE20K~\cite{ade20k} semantic segmentation task ($42.8\%$ mIoU) and LaSOT~\cite{Fan2019LaSOT} visual tracking task ($66.1\%$ AUC) in the lightweight regime. In specific, our pre-trained ViT-Tiny not only outperforms the image classification-proficient LeViT~\cite{levit}, but also achieves comparable accuracy with the current SOTA segmentation-specific models (\ie, SeaFormer-L~\cite{seaformer} and EfficientViT-B1~\cite{efficientvit-iccv}) on ADE20K and even enhances the lightweight OSTrack~\cite{ostrack} to outperform all the current SOTA lightweight CPU-realtime trackers with a new record of $66.1\%$ AUC on LaSOT. 
Our improved code and raw results are released at \hyperlink{https://github.com/wangsr126/mae-lite}{https://github.com/wangsr126/mae-lite}.

This paper extends our conference version published in ICML’23~\cite{mae-lite}, which is the first to successfully perform MIM pre-training for lightweight ViT-Tiny (5.7M) with vanilla design, and has experimentally demonstrated that proper pre-training can bridge the performance gap between vanilla ViT architectures and delicately designed ones in the lightweight regime to a great extent.
The present work goes further to make the following additional contributions in this journal version. (1) We investigate additional pre-training baselines (\ie, CL-based DINO and MIM-based BEiT, BootMAE, MaskFeat, SimMIM) in our study to make our observation-analysis-solution flow more conclusive. (2) We conduct the new experimental study on the ADE20K semantic segmentation and LaSOT visual tracking tasks (see Tabs.~\ref{tab:ade20ksota} and~\ref{tab:lasotsota}), which show that our pre-training enables the more versatile ViT model ViT-Tiny to be outstanding on both of these two tasks. (3) We apply our distillation strategies on the MAE pre-training of the recently proposed Hiera-Tiny (6.5M) to demonstrate the generalization capability of our observation-analysis-solution flow when extended to broader architecture designs. (4) We further improve the original distillation strategy (D-MAE) by solving the issue that the attention weights on the final layer of the obtained ViT-Tiny based on D-MAE exhibit some weird behavior patterns (see~\cref{fig:cmp-distill}). The new decoupled distillation strategy D2-MAE not only consistently improves over D-MAE on all our evaluated datasets (see~\cref{tbl:transfer-distill}), but also enables the pure ViT-Tiny to achieve $79.4\%$ top-1 accuracy on ImageNet-1K, which is comparable to the current lightweight and efficient SOTA networks with sophisticated architecture design, \eg, LeViT-192~\cite{levit}, FastViT-T12~\cite{fastvit}, EfficientFormerV2-S1~\cite{efficientformerv2}, SeaFormer-L~\cite{seaformer}, EfficientViT-B1~\cite{efficientvit-iccv}, \etc (5) More competitors are added for comparison, and many new references are included and summarized in the related works. (6) All  parts of the paper are re-organized and re-written to include the above contributions better.

\section{Related Work}\label{sec:relatedwork}

\subsection{Self-Supervised Learning}

Self-supervised learning (SSL) has received considerable attention recently where a shared good representation is pre-trained for different subsequent downstream tasks without heavy reliance on expensive manual annotations. The richer learning signal in SSL pre-training enables such prior representation learning results to enjoy better transferability to various downstream tasks. This can be achieved by solving some well-designed pretext tasks, such as the pioneering works that adopt rotation prediction~\cite{rotation_pred}, image colorization~\cite{image_colorization}, jigsaw puzzle solving~\cite{jigsaw}, exemplar discrimination~\cite{instance_discrimination}, image inpainting~\cite{pathak2016context} and masked image modeling~\cite{beit}. Some works also exploit automatic grouping of instances in the form of clustering~\cite{Dino2, Dino8, Dino9, Dino36, Dino42, Dino74, Dino80, Dino85}.

\emph{Contrastive learning (CL)}, as one of the discriminative approachs that follow the pioneering exemplar discrimination work, has been very popular in recent years and also achieved incredible success in improving the pre-training of various ConvNets~\cite{simclr, mocov1, mocov2, BYOL, swav, simsiam, Dino23, Dino81} and ViTs~\cite{mocov3, dino}. Concretely, by defining different loss functions, CL pre-training aims to push the representations of different views of the same image (positive pairs) towards each other in the embedding space while the representations of views from different images (negative pairs) are pulled against each other, or even without explicitly discriminating between negative pairs from different images. It relies on different data augmentation strategies to augment the different views. Several other earlier works also explore to match the bag-of-visual-words representations~\cite{Dino26} or train features to align to the targets that are sampled from an uninformative noise distribution~\cite{Dino6}. Recently, some works~\cite{SEED, CompRess, oss} have sought to apply CL on lightweight ConvNets and improve the performance by distillation.

\emph{Masked image modeling (MIM)} with a broad spectrum of design choices, from framework designs~\cite{mae, simmim, splitmask, cae, bootmae} to mask strategy~\cite{hpm, attmask, adios, semmae, um-mae} and prediction target~\cite{beit, ibot, data2vec, peco, milan, maskfeat} designs, has drawn much attention of the vision community. It takes a similar formulation to masked language modeling (MLM) in BERT~\cite{bert}, \ie, learning by recovering the masked patches/tokens from the visible ones in the corrupted image. For instance, BEiT~\cite{beit} is the pioneer along this line which predicts masked tokens generated by a discrete VAE~\cite{dvae} pre-trained on DALLE~\cite{dalle}. MAE~\cite{mae}, SimMIM~\cite{simmim}, and BootMAE~\cite{bootmae} all use RGB pixels as the reconstruction target for groundtruth while they adopt different special designs on the architecture of prediction head. For better supervisions, iBot~\cite{ibot}, data2vec~\cite{data2vec}, and BootMAE~\cite{bootmae} innovatively reconstruct the masked tokens with the online tokenizer outputs of EMA (exponential moving average) updated models as supervision; MaskFeat~\cite{maskfeat} reveals that simple histogram of oriented gradients as in HOG descriptor is a particularly effective target for continuous feature regression; the pre-trained teachers obtained from DINO~\cite{dino} and CLIP~\cite{clip} are also exploited in some works~\cite{hpm, milan, beitv2, maskfeat}. It has been demonstrated that these methods can scale up well on large ViTs, while their performance on lightweight ones is seldom investigated. Some works~\cite{convnextv2, beit, simmim, mcmae} also explore the alternative MIM frameworks for use with ConvNets.

In this paper, we develop and carefully adapt various recently popular MIM pre-training methods to the lightweight regime of ViTs, and also comprehensively benchmark them along with two typical CL pre-training methods and the fully-supervised pre-training methods as the baselines, so as to explore strong lightweight ViTs through MIM pre-training.  

\subsection{Vision Transformers}
Vision transformers (ViTs)~\cite{vit} is the pioneering work that applies a transformer architecture (a stack of attention modules~\cite{attention}) on image patches and shows very competitive results by introducing mean color as prediction targets. Without using any convolution to have an inductive bias\footnote{Different inductive biases can be introduced through architectural choices, the objective function, the curriculum, or the optimization regime.}~\cite{DeiT-1} to focus on spatially neighbor image elements/tokens, this preliminary attempt requires a large and curated dataset to regularize the training and make the learned transformer effective. Recently, the performance of ViTs has been largely improved thanks to better training recipes~\cite{deit, steiner2021train, touvron2022deit}. For instance, DeiT in~\cite{deit} introduces a ConvNet teacher to help the supervised training of the ViT student with vanilla architecture to acquire some inductive biases. Besides, better architecture design is explored in some ViT derivatives~\cite{xcit, cait} which focus on improving self-attention mechanism to boost performance. For instance, CaiT~\cite{cait} introduces the LayerScale and class-attention designs for going deeper with ViTs while XCiT~\cite{xcit} replaces the self attention with a novel cross-covariance attention for reducing the quadratic complexity of self attention to linear complexity.

\emph{Hybrid architecture design} has gained popularity recently with the emergence of ViTs because it can explicitly incorporate necessary inductive biases for ViT-based architectures. ConvNets have benefited from years of manual (\eg,~\cite{Simonyan2015VGG, inception, resnet, densenet, mobilenetv2, shufflenet, tinynet, regnet, convnext, convnextv2}) and automated (\eg,~\cite{efficientnet, mobilenetv3, RegNet14, RegNet15, RegNet17}) network designs, with some of them focusing on finding a good trade-off between computational efficiency (\ie, fewer parameters and lower latency) and accuracy, \eg, the families of RegNet~\cite{regnet}, EfficientNet~\cite{efficientnet}, MobileNet~\cite{mobilenetv2, mobilenetv3}, ConvNeXt~\cite{convnext, convnextv2}, \etc In specific, the ConvNeXt family~\cite{convnext, convnextv2} modernizes the ResNet~\cite{resnet} architecture following the ViT design choices. Inspired by the success of these ConvNets, hybrid architecture design for various ViT derivatives explores to restrict the attention region to nearby tokens~\cite{swin, cswin} or aggregate the neighboring tokens for progressively reduced token numbers~\cite{pvt, t2t-vit, pit}. Many other hybrid works~\cite{botnet, convit, cvt, coat, vit_c, levit, mobilevit, mobilevitv2, mobilevitv3, edgevit, mobileformer, efficientformerv2, edgenext, fastvit} are also inspired by investigating the combination of ConvNets and ViTs. The merit behind this integration is that it can capture both long range dependencies using self-attention mechanism of ViTs and local information using local kernels in ConvNets, and then make a fusion to improve performance on vision tasks.

\emph{Lightweight models} are friendly to resource-constrained devices, so most of the above ViT derivatives have either provided their corresponding lightweight versions or been specifically designed for a better efficiency-accuracy trade-off. For example, the lightweight regime of Mobile-Former~\cite{mobileformer} covers the computational cost range from 26 MFLOPs to 508 MFLOPs. For LeViT~\cite{levit}, it covers the parameter number range from 7.8 M to 39.1 M. Although Mobile-Former has relatively fewer network parameters and costs significantly fewer FLOPs than LeViT, the latter enjoys much higher inference speed thanks to the ConvNet's clothing design without using too many complicated operators. By getting rid of the complex trade-offs among model size (or parameter number), FLOPs, inference speed (or latency), and accuracy of the above ViT derivatives with delicate design, a preliminary version of this paper published in ICML’23~\cite{mae-lite} is the first to systematically study the effects of applying SSL pre-training on lightweight ViTs with vanilla design. It is demonstrated that proper pre-training based on the proposed distillation strategy can even close the performance gap between vanilla ViT architectures and delicately designed ones in the lightweight regime to a great extent.
The present work goes further to investigate more typical pre-training methods to provide more comprehensive analyses, propose an improved decoupled distillation strategy to unleash the great potential of pre-training, and provide more experimental results to make our study more conclusive. The new results on the downstream semantic segmentation and visual tracking tasks are also impressive.

\subsection{Knowledge Distillation}
Knowledge distillation (KD), a mainstream approach for model compression~\cite{compression, hinton2015distilling, fitnets}, is best known as an effective method for taking advantage of a large number of parameters in a teacher model during training, while having an optimized smaller and efficient student model during inference. Beyond model compression, KD is also reckoned to be able to combine strengths of different learning algorithms~\cite{DeiT-1} by transferring the effect of inductive biases between different models. It is claimed that having the right inductive biases is essential for obtaining high performance when data or compute is a limiting factor. For instance, the distillation-based supervised training (or fine-tuning) in DeiT~\cite{deit} achieves better accuracy on vanilla ViTs by adopting a ConvNet as the teacher. Extensive methodologies have also been proposed to improve the distilling effectiveness. In~\cite{hinton2015distilling}, the mimic error of a small student network towards the soft logit output of a larger teacher network is adopted to guide the transferring of the knowledge. FitNet~\cite{fitnets} proposes to distill semantic information from intermediate features, and a MSE-based hint loss is adopted. Besides, the channel attention~\cite{zagoruyko2016at} and the relation of examples~\cite{RelationKD} are considered as the learned knowledge to be transferred. Moreover, KD is widely adopted in various tasks apart from classification tasks, \eg, object detection \cite{KD_Det} and semantic segmentation \cite{KD_Seg}.

\emph{Under self-supervised settings}, there have been a few works performing KD technologies to transfer knowledge from a larger pre-trained teacher to a lightweight student model for better representations, because the naive SSL methods do not work well for lightweight models as shown in~\cite{SEED}. For instance, some works~\cite{distillbert, tinybert, minilmv2, mobilebert} attend to distill large-scale pre-trained language models. In the area of computer vision, CompRess~\cite{CompRess} and SEED~\cite{SEED} employ knowledge distillation to improve the self-supervised visual representation learning capability of small ConvNet models, relying on a CL-based MoCo~\cite{mocov2} framework. A more flexible framework based on CL is proposed in DisCo~\cite{DisCo} to transfer ConvNet teachers' knowledge to smaller students. In~\cite{TinyMIM-46}, a simple feature distillation method is proposed to improve the fine-tuning performance of ViTs based on various pre-training methods, including CL-based approaches such as DINO~\cite{dino} and EsViT~\cite{esvit}, visual-language pre-trained models such as CLIP~\cite{clip}, and supervised models such as DeiT~\cite{deit}. Some methods also adopt online distillation \cite{digo, oss}, which recommends training the teachers and students at the same time and conducting SSL and KD simultaneously. Others focus on taking some inspiration from SSL to boost the performance of task-specific distillation under a fully supervised setting. For example, CRD~\cite{CRD} introduces a new contrastive loss between teacher and student networks for better knowledge distillation. In SSKD~\cite{SSKD}, SSL is treated as an auxiliary task to help extract richer knowledge from a teacher network.

In this paper, we focus on further improving the performance of MIM pre-trained lightweight ViTs by distillation, which is developed following the guidance of our thorough analyses. Our MIM pre-training with distillation not only helps the pre-trained lightweight ViTs to learn semantics at an abstract level relevant to recognition in higher layers, but also makes the downstream models preserve the useful locality inductive bias benefiting from MIM pre-training. 

\section{Observation}
\label{sec:observation}
In this section, we carefully adapt various typical pre-training methods to the lightweight regime of pre-training ViTs, benchmark the pre-trained lightweight models on ImageNet, and further evaluate their transferability to other datasets and tasks, which addresses a little-explored question: how well does pre-training work on lightweight ViTs with vanilla architecture design, especially for MIM pre-training?

\subsection{Preliminaries and Experimental Study Design}
\label{Sec:Preliminaries}
\noindent\textbf{Experimental Unit.} We use a slightly modified version of~\cite{deit} with vanilla lightweight architecture, namely ViT-Tiny, as the experimental unit in our study, which contains 5.7M parameters. We examine the effect of different pre-training methods on it by benchmarking the corresponding downstream performance. Following~\cite{deit}, it consists of one patch embedding layer and twelve transformer blocks with the embedding dimension set to 192. The fixed-size input RGB image at resolution $224\times 224$ is decomposed into a batch of patches of a fixed size of $16\times 16$ pixels. \emph{We slightly modify the number of heads from 3 to 12 as we find this can improve the model's expressive power.}  We choose ViT-Tiny for study because its non-hybrid vanilla architecture has no strong inductive biases by design and its simplicity\footnote{In this section, we also do not use some recent techniques like relative position bias~\cite{swin, levit, Swin49, Swin1, Swin32, Swin33, levit34} and LayerScale~\cite{cait}.} allows us to focus more directly on the impact of pre-training. In addition, ViT-Tiny is an ideal experimental unit on which almost all existing pre-training methods can be directly applied.

\vspace{1.5mm}
\noindent\textbf{Evaluation Protocols.}
\emph{Linear probing} has been a standard evaluation protocol to compare the quality of the pre-trained weights~\cite{mocov1, mocov2, BYOL, swav}, in which only the prediction head is tuned based on the downstream training set while the pre-trained representations are kept frozen. However, prior works point out that linear evaluation does not always correlate with the practical utility~\cite{mae,newell2020useful}. 
In this study, we mainly adopt \emph{fine-tuning} as the default evaluation protocol to benchmark the downstream performance, in which all the layers are tuned by initializing them with the pre-trained models. We leave the \emph{linear probing} evaluation for analysis in~\cref{Sec:Work} to investigate the secrets behind pre-training.
By default, we do the \emph{fine-tuning} evaluation on ImageNet~\cite{ImageNet} by tuning the whole models on its training set and then evaluating them on the validation set. Several other downstream classification tasks, object detection and segmentation tasks, semantic segmentation task, and single object tracking task are also exploited for comparison.

\vspace{1.5mm}
\noindent\textbf{Compared Methods.} We develop and carefully adapt various recently popular MIM pre-training methods (mainly for the large models) to apply them on ViT-Tiny, \ie, the MAE~\cite{mae} and SimMIM~\cite{simmim} methods which aim to predict raw RGB pixels of masked patches, and some methods with other pre-defined targets for prediction (\eg, discrete tokens in BEiT~\cite{beit}, features from a momentum teacher in BootMAE~[2], and HoG features in MaskFeat~[72]). We also benchmark them along with two CL pre-training methods (\ie, MoCo-v3~\cite{mocov3} and DINO~\cite{dino}) and the fully-supervised pre-training methods as the baselines.

\emph{Baseline without pre-training.} Largely following the recipe in~\cite{deit}, we train a ViT-Tiny model from scratch for 300 epochs on the training set of ImageNet-1K (dubbed IN1K), which serves as the baseline without pre-training in our study. 
It achieves 74.5\% top-1 accuracy on the validation set of IN1K, surpassing that achieved based on the original architecture with 3 heads (\ie, 72.2\% in~\cite{deit}). It further reaches 75.8\% by adopting our improved training recipe with some optimized hyper-parameters for augmentations (\ie, weaker regularization and augmentation than training large-scale models). Besides, layer-wise \emph{lr} decay is also taken into consideration during \emph{fine-tuning} following~\cite{beit, mae}. Please refer to Appendix~\ref{Appdix:appdix-eval-detail} for more details. This can be seen as a strong baseline to examine the pre-training in our study because it even surpasses the original transformer-specific distillation-based version in~\cite{deit} (\ie, 74.5\%).

\emph{Supervised pre-training baselines.} In~\cite{steiner2021train}, a comprehensive study of the trade-offs between model regularization, data augmentation, training data size and compute budget in ViTs is conducted. It is evident that the increased compute and AugReg (\ie, a combination of modern data augmentation and model regularization techniques) can yield significantly better performance than the plain pre-training on the training set of ImageNet-21K (dubbed IN21K). So we follow the supervised pre-training recipe in~\cite{steiner2021train} and scale up the pre-training duration on IN21K from 30 to 300 epochs to serve as our supervised pre-training baselines.

\emph{CL pre-training baselines.} To adapt MoCo-v3~\cite{mocov3} and DINO~\cite{dino} to ViT-Tiny, we re-implement them with ViT-Tiny as encoder and largely follow the original setups in their papers. We conduct 400-epoch pre-training on IN1K for them. In~\cite{mocov3}, it is claimed that instability is a major issue that impacts self-supervised ViT pre-training and causes mild degradation in accuracy, and a simple trick by adopting fixed random patch projection (the first layer of a ViT model) is proposed to improve the stability in practice. However, we find that instability is not the main issue for MoCo-v3 adaptation to ViT-Tiny, and higher performance is achieved with a learned patch projection layer. In~\cite{dino}, DINO is interpreted as a form of knowledge \textbf{di}stillation with \textbf{no} labels, where a centering and sharpening of the teacher output is used to avoid collapse.

\emph{MIM pre-training methods.} We carefully adapt MAE~\cite{mae}, BEiT~\cite{beit}, BootMAE~\cite{bootmae} and MaskFeat~\cite{maskfeat} to ViT-Tiny by investigating the different effects of several basic factors and components in their frameworks. For instance, as a key component, decoder determines the semantic level of the learned latent representations. Although MIM works well across a wide range of the decoder’s width and depth under the \emph{fine-tuning protocol} when the encoder is large, we experimentally find that MIM prefers a much more lightweight decoder when the encoder is altered to ViT-Tiny, \ie, a decoder consisting of one transformer block with the head number and width set to 3 and 96. We also conduct a small sweep over 5 masking ratios $\{0.45, 0.55, 0.65, 0.75, 0.85\}$ and find that 0.75 achieves the best performance. Other setups in the above papers remain intact including the optimizer, learning rate, batch size, argumentation, \etc 
As for SimMIM~\cite{simmim} adaptation to ViT-Tiny, we rigorously follow the original settings in the paper.
We also do MIM pre-training on IN1K for 400 epochs.

\subsection{Fine-Tuning Evaluation on ImageNet Classification}
\label{Sec:Benchmarks}

\noindent\textbf{\emph{Fine-Tuning} Evaluation.} We follow the practice of supervised ViT-Tiny training with our improved recipe in Appendix~\ref{Appdix:appdix-eval-detail} for \emph{fine-tuning} evaluation, though the decay coefficient in the layer-wise \emph{lr} decay schedule is separately optimized for different pre-trained models. Besides, we use global average pooling (GAP) after the final block during the \emph{fine-tuning} of both the MIM and CL pre-trained models, which is, however, not the common practice for MoCo-v3~\cite{mocov3}. We adopt GAP for MoCo-v3 as it significantly helps to surpass the model using the original configuration based on a class token (76.8\% \vs 73.7\%) for our used lightweight ViT-Tiny.

\begin{table*}[!t]
	\begin{center}
        \setlength{\tabcolsep}{2.0pt}
		\caption{\textbf{Comparison across different pre-training methods by evaluating the pre-trained models on data-sufficient IN1K.} We report top-1 accuracy on its validation set. Our modified ViT-Tiny with vanilla architecture is used as the experimental unit. The pre-training time is measured on an 8$\times$V100 GPU machine. `Ori.' represents the original training/fine-tuning recipe in DeiT~\cite{deit} and `Ours' represents our improved recipe (see Appendix~\ref{Appdix:appdix-eval-detail}).}
        \fontsize{8pt}{12pt}\selectfont
		\begin{threeparttable}
		\begin{tabular}{llccccccc}
			
			\toprule
			\multirow{2}{*}{\textbf{Models}}&\multicolumn{4}{c}{\textit{Pre-Training}}&\multicolumn{3}{c}{\textit{Fine-Tuning}}&\textbf{Top-1 }\cr
			\cmidrule(lr){2-5} \cmidrule(lr){6-8}
			&\textbf{Methods}&\textbf{Data} &\textbf{\#Epochs}&\textbf{Time(h)}&\textbf{Recipe}&\textbf{Distill.}&\textbf{\#Epochs}& (\%)\cr
			\cmidrule(lr){1-9}
			&\colorgray{\textit{MIM Self-Supervised}}&&&&&&&\cr
			\multirow{16}{*}{ViT-Tiny ($5.7M$)}&MAE~\cite{mae}&IN1K w/o labels&400&23&Ours&\XSolidBrush&300&78.0 \cr
   
			&SimMIM~\cite{simmim}&IN1K w/o labels&400&40&Ours&\XSolidBrush&300&77.9 \cr
   
			&BEiT~\cite{beit}&IN1K w/o labels&400&59&Ours&\XSolidBrush&300&77.7 \cr
   
			&BootMAE~\cite{bootmae}&IN1K w/o labels&400&30&Ours&\XSolidBrush&300&77.9\cr
   
			&MaskFeat~\cite{maskfeat}&IN1K w/o labels&400&34&Ours&\XSolidBrush&300&77.9\vspace*{1pt}\cr

            &\colorgray{\textit{CL Self-Supervised}}&&&&&&&\cr
            
			\colorgray{\textit{Vanilla Architecture}}&DINO~\cite{dino}&IN1K w/o labels&400&83&Ours&\XSolidBrush&300&77.2\cr
   
			&MoCo-v3~\cite{mocov3}&IN1K w/o labels&400&52&Ours&\XSolidBrush&300&76.8\vspace*{1pt}\cr

            &\colorgray{\textit{Supervised}}&&&&&&&\cr
   
			&AugReg~\cite{steiner2021train}&IN21K w/ labels&300&200&Ours&\XSolidBrush&300&77.8 \cr
   
			&AugReg~\cite{steiner2021train}&IN21K w/ labels&30&20&Ours&\XSolidBrush&300&76.9 \cr

            &Our Recipe&IN1K w/ labels&300&20&Ours&\XSolidBrush&300&76.5\vspace*{1pt}\cr

            &\colorgray{\textit{W/O Pre-Training}}\\
			&Random Init.~\cite{deit}&-&-&0&Ours&\XSolidBrush&300&75.8 \cr

            &Random Init.~\cite{deit}&-&-&0&Ori.&\XSolidBrush&300&74.5 \cr

			\bottomrule
		\end{tabular}
      \end{threeparttable}
		\label{tab:imagenetcompare}
    \vspace{-0pt}
	\end{center}
\end{table*}

\vspace{1.5mm}
\noindent\textbf{Observation 1.} As shown in~\cref{tab:imagenetcompare}, almost all of the compared pre-training methods, including both the fully-supervised and self-supervised ones, can achieve better downstream fine-tuning performance on the ImageNet classification task than the random initialization paradigm without pre-training, whilst \emph{MIM pre-training with moderate pre-training cost consumed outperforms the compared pre-training baselines}. As we scale up the supervised pre-training duration on IN21K from 30 to 300 epochs, we find that the $10\times$ longer schedule can achieve a comparable performance (\ie, $77.8\%$) to the MIM pre-training, while the pre-training cost gets much higher due to the significantly increased pre-training time, let alone the used expensive labels. We also conduct the supervised pre-training with 300 epochs on IN1K using our fine-tuning recipe to better show the benefits of self-supervised pre-training or increasing the training data size.
All the results indicate that the \emph{extremely simple} lightweight ViTs with vanilla design have great potential which can be unleashed via proper pre-training.

\begin{table*}[h!]
\setlength{\tabcolsep}{11pt}
\begin{center}
\renewcommand{\arraystretch}{0.8} 
\caption{\textbf{Effect of varying the SSL pre-training data size and class distribution based on ViT-Tiny}. We follow the \emph{400-epcho self-supervised pre-train, 300-epoch supervised fine-tune} paradigm. Top-1 accuracy on the validation set of IN1K is reported. 
}
\label{tbl:pretrain-dataset}
\fontsize{8pt}{12pt}\selectfont
\begin{tabular}{cccccc}
\toprule
\diagbox{\textbf{Methods}}{\textbf{Datasets}} & IN1K 100\% & IN1K 10\% & IN1K 1\% & IN1K-LT & IN21K \\
\midrule
MoCo-v3~\cite{mocov3} & 76.8 & 76.5 \small\colorred{(-0.3)} & 76.2 \small\colorred{(-0.6)} & 76.1 \small\colorred{(-0.7)} & 76.9 \small\colorgreen{(+0.1)} \\
MAE~\cite{mae} & 78.0 & 78.0 \small\colorgreen{(+0.0)} & 77.9 \small\colorred{(-0.1)} & 77.9 \small\colorred{(-0.1)} & 78.0 \small\colorgreen{(+0.0)} \\
\bottomrule
\end{tabular}
\end{center}

    \vspace{-0pt}
\end{table*}

\vspace{1.5mm}
\noindent\textbf{Observation 2.} The above observation encourages us to further explore how the enhanced lightweight ViTs perform in comparison to the recent SOTA lightweight ViT derivatives (including both the hybrid and non-hybrid) as introduced in~\cref{tab:sota}. It can be observed that \emph{the enhanced vanilla ViT-Tiny through MIM pre-training can also achieve SOTA performance on par with some superior ViT derivatives with comparable parameter number.} It is noteworthy that these delicately designed architectures in the lightweight regime, despite lack of pre-training, achieve top performance on the ImageNet classification task by seeking to (1) extend the supervised fine-tuning duration to a much longer schedule (\eg, 1000 epochs in LeViT~\cite{levit} and PiT~\cite{pit}), (2) introduce distillation-driven training to regularize the fine-tuning (\eg, LeViT~\cite{levit}, XCiT~\cite{xcit}, PiT~\cite{pit}, and CaiT~\cite{cait}, \etc), or (3) use some recent techniques like relative position bias~\cite{Swin49, Swin1, Swin32, Swin33, levit34} (\eg, LeViT~\cite{levit} and Swin~\cite{swin}), conditional positional encoding~\cite{EdgeViT8} (\eg, EdgeViT~\cite{edgevit}) and LayerScale (\eg, XCiT~\cite{xcit}). The competing results of the enhanced ViT-Tiny demonstrate the benefits of studying MIM pre-training on \emph{extremely simple} lightweight ViTs, which is orthogonal to the network architecture design methodology. Besides, \emph{extremely simple} architectures are usually friendly to the real-world deployment by getting rid of the complicated modules existing in some ViT derivatives (\eg, EdgeViT-XS~\cite{edgevit}, MobileViT-S~\cite{mobilevit}, Mobile-Former-294M~\cite{mobileformer}, SeaFormer-L~\cite{seaformer}).  

\vspace{1.5mm}
\noindent\textbf{Observation 3.} 
We also empirically examine whether the self-supervised pre-training on lightweight ViTs can benefit from large-scale pre-training data.
In~\cref{tbl:pretrain-dataset}, we study the effect of varying the pre-training data size and class distribution with two typical SSL methods (\ie, MoCo-v3 and MAE) applied on ViT-Tiny. 
We first consider the much larger dataset IN21K, however, few improvements are observed for both MoCo-v3 and MAE. We believe the limited capacity of \emph{extremely simple} lightweight ViTs also bottlenecks the representation quality of self-supervised pre-training even with the ``LLM-like” scaling of data. We further consider two subsets of IN1K containing 10\% and 1\% of the total examples (IN1K 10\% and IN1K 1\%) balanced in terms of classes~\cite{assran2021semi} and one subset with long-tailed class distribution~\cite{imagenet-lt} (IN1K-LT). Surprisingly, MAE pre-training has marginal performance declines on these subsets, showing more robustness than MoCo-v3 in terms of the pre-training data size and class distribution.

\subsection{Transfer Learning Evaluation on Downstream Tasks}
\label{Sec:Transfer}
We further examine the transfer learning performance of the pre-trained models in~\cref{tbl:transfer}. We only consider the models pre-trained based on IN1K for simplicity, involving their transfer performance on some other downstream classification tasks and dense prediction tasks.

\begin{table*}[h!]
\setlength{\tabcolsep}{0.5pt}
\begin{center}
\renewcommand{\arraystretch}{0.9} 
{
\caption{\textbf{Transfer evaluation of our modified ViT-Tiny on some data-insufficient downstream classification tasks and dense prediction tasks}. Top-1 accuracy is reported for classification tasks, mIoU is for ADE20K segmentation, AUC is for LaSOT tracking, and AP is for COCO detection and segmentation (Det./Seg.).
The dataset scale is represented as (Train-size per epoch/\#Train-classes).}
\label{tbl:transfer}
\fontsize{8pt}{12pt}\selectfont
\begin{tabular}{cccccccccc}
\toprule
\multirow{3}{*}{\diagbox{\textbf{Methods}}{\textbf{Datasets}}} & \multicolumn{6}{c}{\textit{Downstream Classification Tasks}}&\multicolumn{3}{c}{\textit{Downstream Dense Prediction Tasks}} \\ \cmidrule(lr){2-7} \cmidrule(lr){8-10}
& \textbf{Flowers} & \textbf{Pets} & \textbf{Aircraft} & \textbf{Cars} & \textbf{CIFAR100} & \textbf{iNat18} & \textbf{ADE20K} & \textbf{LaSOT} & \textbf{COCO}\\
& \scriptsize{(2k/102)} & \scriptsize{(4k/37)} & \scriptsize{(7k/100)} & \scriptsize{(8k/196)} & \scriptsize{(50k/100)} & \scriptsize{(438k/8142)} & \scriptsize{(20k/150)} & \scriptsize{(60k/201)} & \scriptsize{(118k/80)}\\
\cmidrule(lr){1-7} \cmidrule(lr){8-10}
W/O Pre-Training & 30.2 & 26.1 & 9.4 & 6.8 & 42.7 & 57.1 & 21.3 & 56.1 & 32.7/28.9\\
\cmidrule(lr){1-7} \cmidrule(lr){8-10}
\colorgray{\textit{Supervised}~(IN1K)}\\
Our Recipe & \textbf{96.4} & \textbf{93.1} & 73.5 & \textbf{85.6} & \textbf{85.8} & \textbf{64.7} & \textbf{41.5} & 64.1 & 40.4/35.5\\
\cmidrule(lr){1-7} \cmidrule(lr){8-10}
\colorgray{\textit{CL Self-Supervised}}\\
DINO~\cite{dino} & 95.6 & 89.3 & 73.6 & 84.5 & 84.7 & \textbf{64.7} & 39.8 & \textbf{64.3} & \textbf{41.4}/\textbf{36.7}\\
MoCo-v3~\cite{mocov3} & 94.8 & 87.8 & \textbf{73.7} & 83.9 & 83.9 & 63.7 & 38.1 & 62.9 & 39.7/35.1\vspace*{3pt}\\
\colorgray{\textit{MIM Self-Supervised}} \\
MaskFeat~\cite{maskfeat} & 92.2 & 85.5 & 72.9 & 82.4 & 82.6 & 64.2 & 38.1 & 63.8 & 41.4/36.5\\
BootMAE~\cite{bootmae} & 89.4 & 79.6 & 69.8 & 79.4 & 80.0 & 63.6 & 36.4 & 63.4 & 39.9/35.4\\
BEiT~\cite{beit} & 89.2 & 78.9 & 65.8 & 77.6 & 79.2 & 63.2 & 35.7 & 62.1 & 39.1/34.7\\
MAE~\cite{mae} & 85.8 & 76.5 & 64.6 & 78.8 & 78.9 & 64.0 & 34.5 & 61.6 & 39.9/35.4\\
SimMIM~\cite{simmim} & 77.2 & 68.9 & 55.9 & 70.4 & 77.7 & 63.8 & 35.5 & 63.4 & 39.3/34.8\\
\bottomrule
\end{tabular}}
\end{center}

\vspace{-0pt}
\end{table*}

\emph{For the classification tasks}, we investigate the performance on the relatively smaller datasets than IN1K to question if the pre-trained models transfer well on these data-insufficient tasks, including Flowers-102~\cite{flower} (Flowers for short), Oxford-IIIT Pets~\cite{oxford-pet} (Pets), FGVC-Aircraft~\cite{aircraft} (Aircraft), Stanford Cars~\cite{stanford-cars} (Cars), CIFAR100~\cite{cifar}, iNaturalist 2018~\cite{inat} (iNat18). For all these datasets except iNat18, we conduct a small sweep for some key fine-tuning parameters, \ie, the learning rate, training epoch number and layer-wise \emph{lr} decay coefficient, to achieve the optimal fine-tuning results on these datasets (see Appendix~\ref{Appdix:appdix-transfer-detail} for more details). Besides, we adopt random resized crop and random horizontal flipping as the augmentations and do not use any regularization (\eg, weight decay, dropout, or the stochastic depth regularization technique~\cite{huang2016deep}). For iNat18, we follow the same training configurations to those on ImageNet (see Appendix~\ref{Appdix:appdix-eval-detail}).

\emph{For the dense prediction tasks}, we mainly consider the object detection and segmentation tasks on COCO~\cite{coco}, semantic segmentation task on ADE20K~\cite{ade20k}, and visual tracking task on LaSOT~\cite{Fan2019LaSOT}. For the tasks on COCO, we reproduce and rigorously follow the setup in~\cite{li2021benchmarking}, except replacing the backbone with ViT-Tiny and decreasing the input image size from 1024 to 640 to make it trainable on a single machine with 8 NVIDIA V100s like other lightweight detectors. We fine-tune for up to 100 epochs with different pre-trained models as the initialization of the backbone. We do not use layer-wise \emph{lr} decay since we find it useless for the tiny backbone on these COCO tasks. The weight decay is 0.05 and the stochastic depth regularization~\cite{huang2016deep} is not used. For the semantic segmentation task, we reproduce the same setup in MAE~\cite{mae} and adopt UperNet framework~\cite{upernet} for evaluation, which follows the semantic segmentation code of~\cite{beit}. In our reproduction, we only replace the backbone encoder with different pre-trained ViT-Tiny models as the initialization and keep the decoder (several deconvolution layers) intact. The input image size is 512$\times$512. For the tracking task, we experiment on the large-scale dataset LaSOT~\cite{Fan2019LaSOT} using a recent strong baseline tracker OSTrack~\cite{ostrack}. We reproduce the setup of speed-oriented version of OSTrack with input image resolution set to 256$\times$256 and only replace the backbone, which is responsible for joint feature extraction and relation modeling, with ViT-Tiny and use different pre-trained ViT-Tiny models as the initialization.

\vspace{1.5mm}
\noindent\textbf{Observation 4.}
 As shown in~\cref{tbl:transfer}, we first find it hard to achieve promising transfer evaluation results for our modified ViT-Tiny on the evaluated downstream tasks by training from scratch without any pre-training. The results significantly lag behind those models that adopt various pre-training. 
 Second, despite that using various pre-training methods shows better performance, the relative superiority and inferiority comparisons between these pre-training methods exhibit distinct characteristics from those on ImageNet (see~\cref{tab:imagenetcompare}).
 We find that \emph{downstream data scale matters} during transfer learning. The MIM-based pre-training approaches tend to achieve downstream performance inferior to the fully-supervised counterpart, while the performance gap is narrowed more or less as the data scale of the downstream task increases. Moreover, MIM-based pre-training even shows inferior results to CL-based approaches. We conjecture that it is due to their different layer behaviors during pre-training and fine-tuning as discussed in detail in the following section.


\section{Analysis}
\label{Sec:Work}
In this section, we introduce \emph{linear probing} evaluation and some model analysis methods to reveal the secrets of pre-training, study the pattern of layer behaviors during pre-training and fine-tuning, and investigate what matters for downstream performances. For the sake of simplicity, we only investigate the ImageNet classification task and other downstream classification tasks shown in~\cref{tbl:transfer} and use the supervised pre-training baseline, the CL-based entries DINO~\cite{dino} and MoCo-v3~\cite{mocov3}, the best MIM-based entry MaskFeat~\cite{maskfeat}, and the worst two MIM-based entries MAE~\cite{mae} and SimMIM~\cite{simmim} in~\cref{tbl:transfer} for analysis in this section.


\begin{table}[t]
\setlength{\tabcolsep}{1.5pt}
\begin{center}

\caption{\textbf{\emph{Linear Probing} Evaluation} of Pre-Trained Models on the Classification Tasks. Top-1 Accuracy Is Reported.
}
\vspace{-2pt}
\label{tbl:appdix-lp}
\fontsize{8pt}{12pt}\selectfont
\begin{tabular}{ccccccc}
\toprule
\multirow{2}{*}{\emph{Methods}} & \multicolumn{6}{c}{\emph{Datasets}} \\
& \textbf{Flowers} & \textbf{Pets} & \textbf{Cars} & \textbf{CIFAR} & \textbf{iNat18} & \textbf{IN1K} \\
 \midrule
 \colorgray{\textit{Supervised}} \\
Our Recipe & 91.0 & \textbf{92.0} & 47.9 & 73.6 & 39.8 & - \\
 \midrule
 \colorgray{\textit{CL}}\\
DINO & \textbf{95.0} & 90.9 & \textbf{53.6} & \textbf{74.9} & \textbf{44.0} & \textbf{66.1} \\
MoCo-v3 & 93.2 & 83.5 & 44.5 & 73.4 & 36.2 & 62.1\vspace*{1pt}\\
\colorgray{\textit{MIM}} \\
MaskFeat & 65.5 & 34.2 & 13.8 & 39.6 & 3.9 & 23.1 \\
MAE & 48.9 & 25.0 & 8.8 & 31.0 & 1.4 & 23.3 \\
SimMIM & 44.1 & 18.5 & 6.6 & 23.3 & 0.9 & 14.8 \\
\toprule
\end{tabular}
\item 
\end{center}
\vspace{-30pt}
\end{table}

\subsection{\emph{Linear Probing} Evaluation}
\label{Sec:lp-analyses}
We first learn linear classifiers on frozen features for different classification tasks in our \emph{linear probing} evaluation. It reflects how the representations obtained by the pre-trained models are linearly separable \wrt semantic categories. As shown in~\cref{tbl:appdix-lp}, the \emph{linear probing} performance is consistently lower than the \emph{fine-tuning} performance except for DINO on Pets. Furthermore, while \emph{linear probing} evaluation is informative for exploiting the pre-trained frozen models, it may not correspond to which models perform best after fine-tuned, especially for those downstream tasks with relatively sufficient labeled data, \eg, iNat18 and IN1K. In other words, it may lead to an underestimation of the value of some pre-trained models in the practical utility~\cite{mae,newell2020useful} on downstream tasks. We attribute it to that \emph{linear probing} only evaluates the final representation of the pre-trained models, making it overlook the value of providing good initialization for lower layers.

\begin{figure*}[thbp!]
    \begin{minipage}[t]{0.26\textwidth}
    \centering
    \begin{tabular}{@{\extracolsep{\fill}}c@{}c@{}c@{}@{\extracolsep{\fill}}}
    \includegraphics[width=0.92\textwidth]{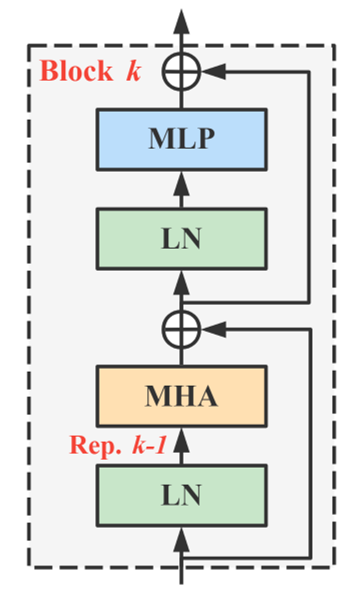}
    \end{tabular}
    \vspace{-0pt}
    \caption{\textbf{Transformer block,} where the feature map after the first LN of block $k$ can be used as normalized output representation for block $k-1$.}
    \label{fig:appdix-transformer}
    \end{minipage}
    \hspace{0.03\textwidth}
     \begin{minipage}[t]{0.68\textwidth}
     \centering
    \begin{tabular}{@{\extracolsep{\fill}}c@{}c@{}c@{}@{\extracolsep{\fill}}}
        \includegraphics[width=0.28\textwidth]{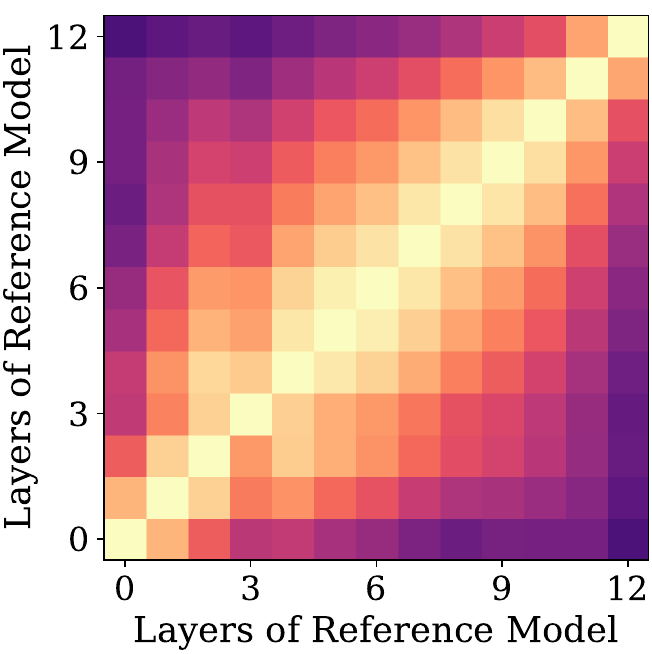}
        \hspace{0.005\textwidth}
        \includegraphics[width=0.28\textwidth]{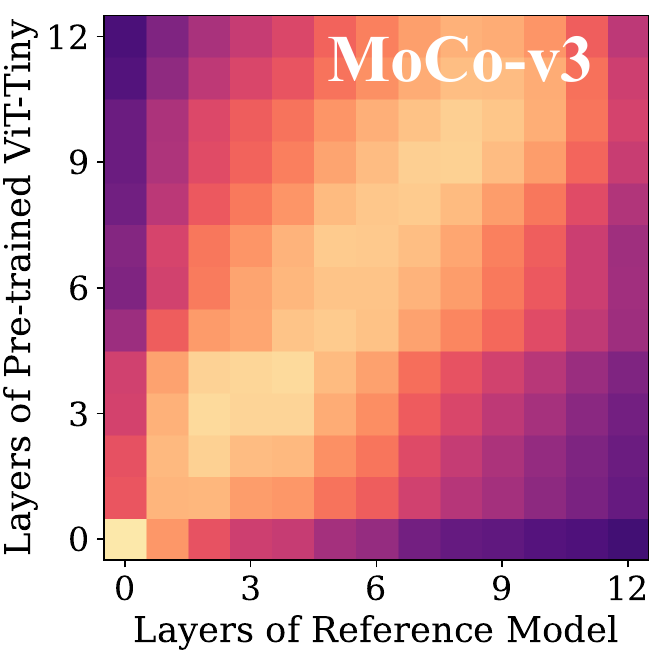}
        \hspace{0.005\textwidth}
        \includegraphics[width=0.34\textwidth]{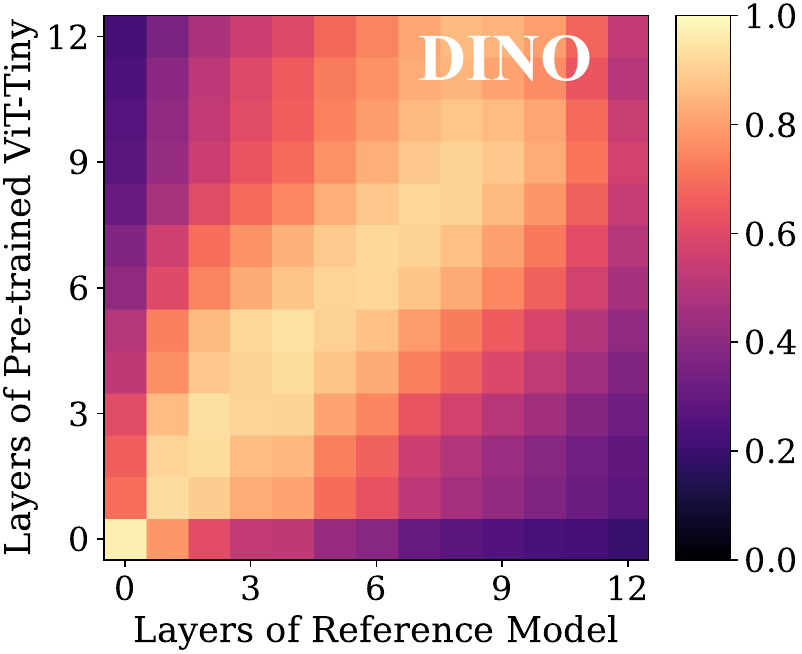} \\
        \includegraphics[width=0.28\textwidth]{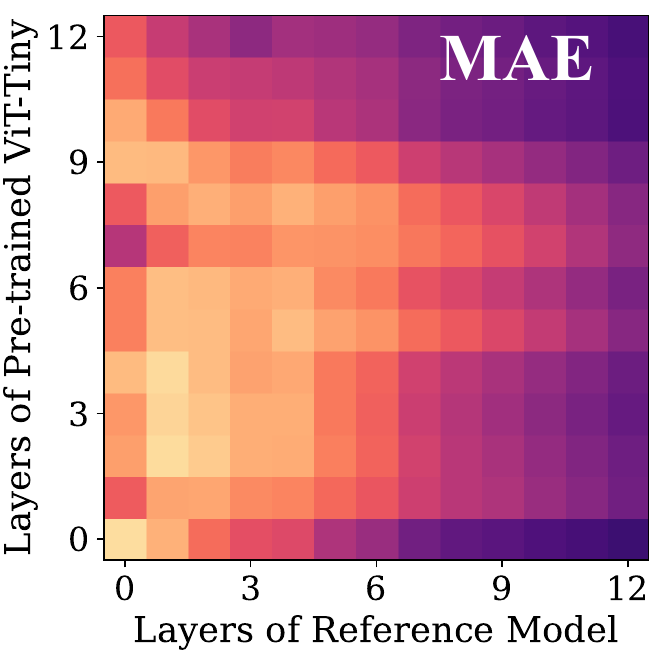}
        \hspace{0.005\textwidth}
        \includegraphics[width=0.28\textwidth]{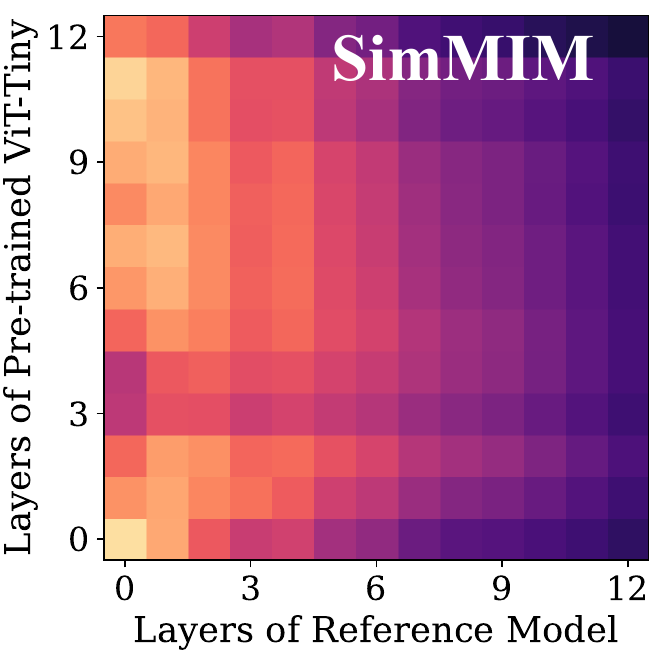}
        \hspace{0.005\textwidth}
        \includegraphics[width=0.34\textwidth]{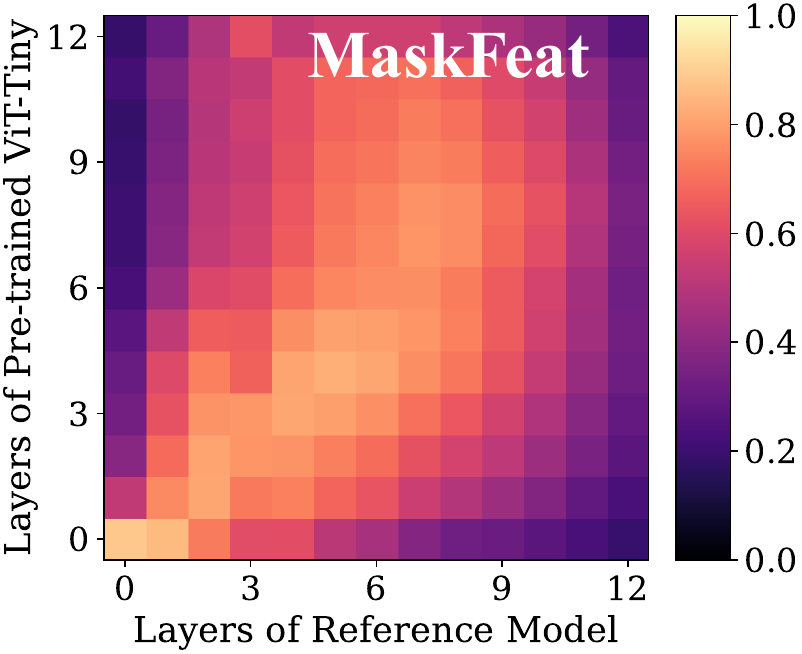} \\
    \end{tabular}
    \vspace{-0pt}
    \caption{\textbf{Layer representation similarity} as heatmaps for the investigated ViT-Tiny models from different pre-training methods, with x and y axes indexing the layers (the 0 index indicates the patch embedding layer), and higher values indicate higher similarity. The fully-supervised baseline based on our recipe (see~\cref{tab:imagenetcompare}) on IN1K is used as the reference.
    }
    \label{fig:cmp}
    \end{minipage}
    \vspace{-0pt}
\end{figure*}

From another point of view, we think the failure of MIM-based self-supervised pre-training to perform better than CL-based models for the data-insufficient downstream tasks (as shown in~\cref{tbl:transfer}) is due to the reason that CL pre-training learns more semantics at an abstract level relevant to recognition in higher layers than MIM pre-training, which also helps the frozen models from CL pre-training to achieve better \emph{linear probing} evaluation results than MIM (as shown in~\cref{tbl:appdix-lp}). This analysis is already evident in the CL-based DINO~\cite{dino} as its pre-training results show that the class-specific features can be automatically learned, even leading to unsupervised object segmentation. This motivates us to further analyze the layer-wise representations and attention maps across and within different models to delve into the working mechanism of the different SSL pre-training methods for lightweight ViTs.

\subsection{Layer-Wise Representation Analysis}
\label{Sec:rep-analyses}
We adopt the Centered Kernel Alignment (CKA) metric~\cite{cortes2012algorithms, nguyen2020wide} to analyze the layer-wise representation similarity across and within different models. Specifically, CKA takes two feature maps (or representations) $\bm{X}$ and $\bm{Y}$ as input and computes their normalized similarity $S(\bm{X},\bm{Y})$ in terms of the Hilbert-Schmidt Independence Criterion (HSIC~\cite{song2012feature}) as
\begin{align}
    S(\bm{X},\bm{Y})&=\mathrm{CKA}(\bm{K},\bm{L}) \nonumber\\ &=\frac{\mathrm{HSIC}(\bm{K},\bm{L})}{\sqrt{\mathrm{HSIC}(\bm{K},\bm{K})\mathrm{HSIC}(\bm{L},\bm{L})}},
\end{align}
where $\bm{K}=\bm{X}\bm{X}^{\top}$ and $\bm{L}=\bm{Y}\bm{Y}^{\top}$ denote the Gram matrices for the two feature maps, and the HSIC computation is invariant to the orthogonal transformation of representations and isotropic scaling. 
In our analyses, a mini-batch version of HSIC is adopted by using an unbiased estimator of HSIC~\cite{nguyen2020wide} to work at scale compatible with our studied models. We select the normalized version of the output representation of each transformer block for computation, \ie, the feature map after the first LayerNorm (LN)~\cite{layer-norm} in its next block (see~\cref{fig:appdix-transformer}).



\vspace{1.5mm}
\noindent\textbf{Analysis 1.} We use heatmaps to visualize the layer-wise representation similarity of investigated pre-trained models with respect to a fully-supervised reference model in~\cref{fig:cmp}. We choose the fully-supervised baseline trained on the IN1K classification task based on our recipe (see~\cref{tab:imagenetcompare}) as the reference because we consider its representations are more recognition aligned and the higher similarity of the examined model's layers to this supervised baseline can indicate more relevance to recognition for the examined model. Although the similarity does not directly indicate whether the downstream performance is good, it indeed reflects the pattern of layer representations to a certain extent. The similarity within the supervised baseline is also presented in~\cref{fig:cmp}. In Appendix~\ref{Appdix:appdix-ref}, we also replace the reference model with some much stronger ViT-Base models to test the robustness of our analysis.

\begin{figure*}[t]
    \centering
    \includegraphics[width=0.235\textwidth]{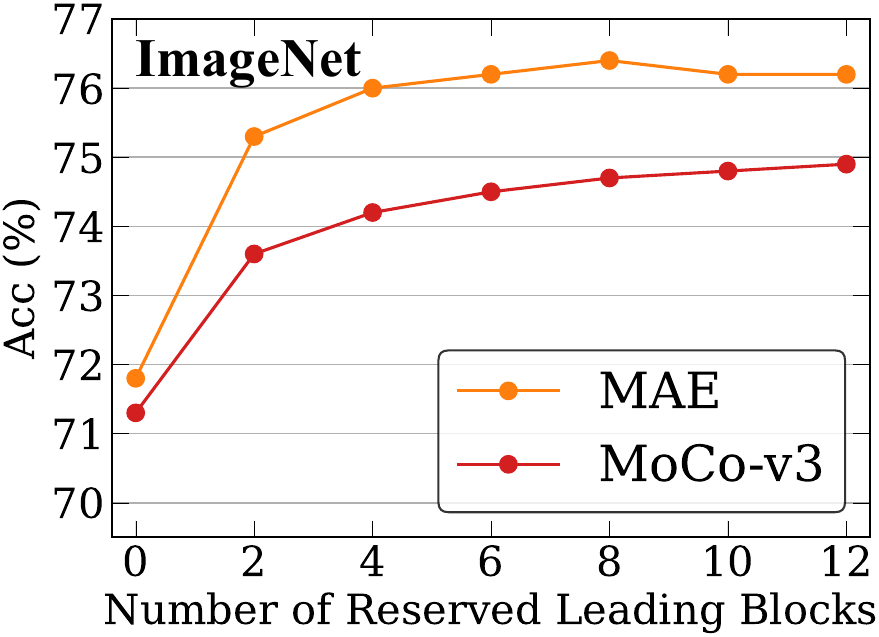}
    \hspace{0.005\textwidth}
    \includegraphics[width=0.235\textwidth]{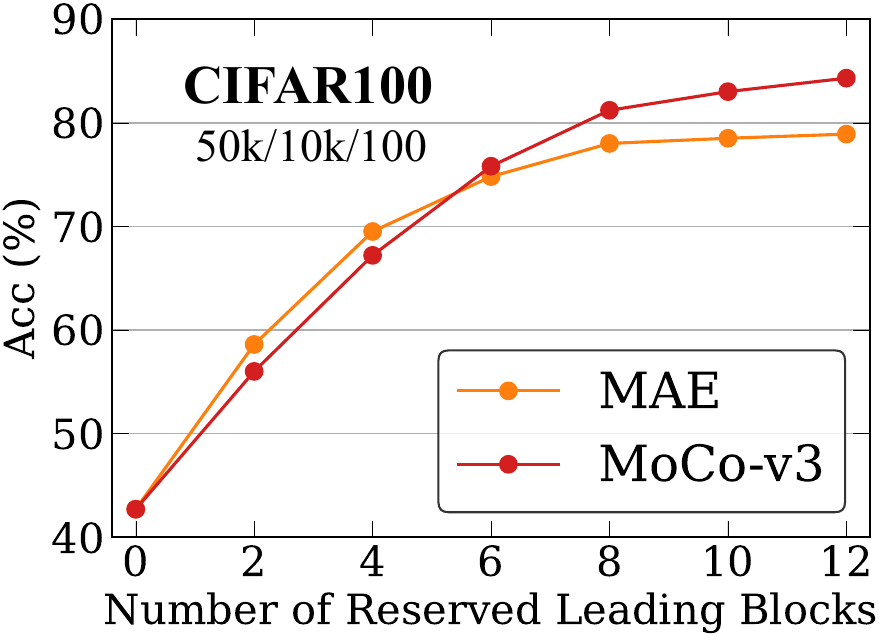}
    \hspace{0.005\textwidth}
    \includegraphics[width=0.23\textwidth]{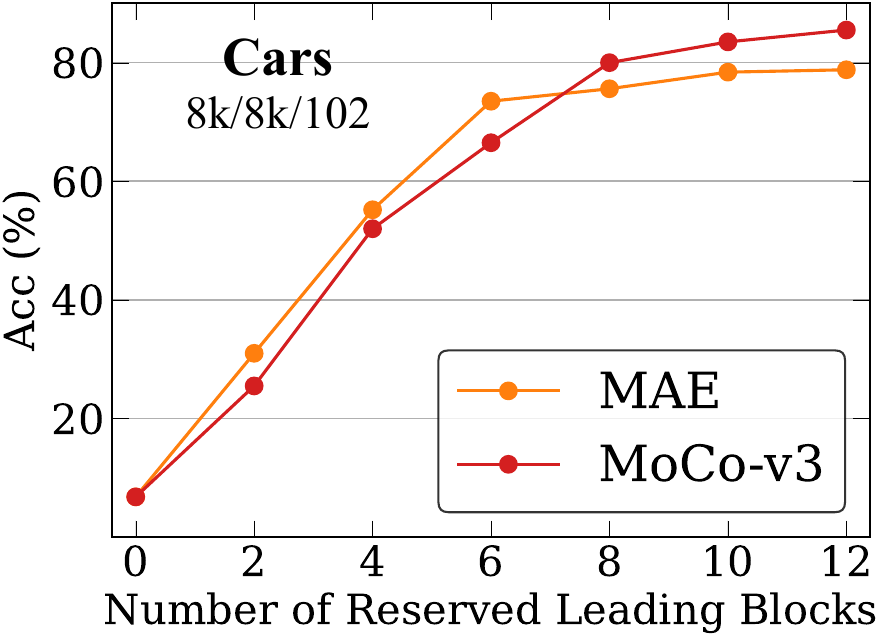}
    \hspace{0.005\textwidth}
    \includegraphics[width=0.23\textwidth]{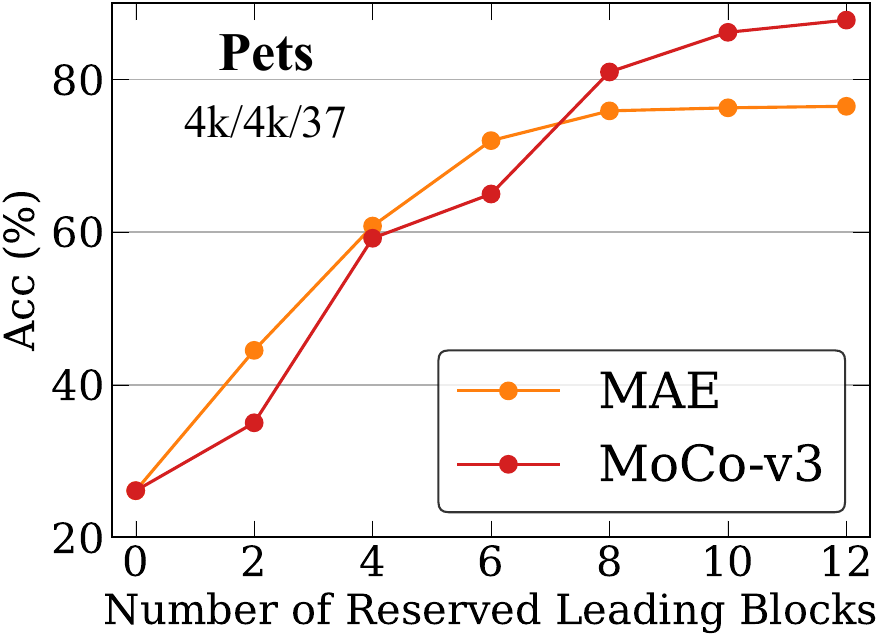}
    \vspace{-0pt}
    \caption{Lower layers of the pre-trained ViT-Tiny models contribute to most gains on the data-sufficient IN1K. The contributions from higher layers of the pre-trained models increase as the downstream dataset scale shrinks, which indicates that higher layers matter in data-insufficient downstream tasks.
    }
    \label{fig:drop-transfer}
    \vspace{-0pt}
\end{figure*}

We notice that lower layers matter more than higher ones if sufficient downstream data is provided. Specifically, we first observe a relatively high similarity between the models from MIM pre-training and the reference model for lower layers, while much lower similarity for higher layers. It indicates that fewer semantics are extracted from MIM pre-training at a more abstract level in higher layers. In contrast, the models from CL pre-training aligns the reference model well across almost all layers. However, the fine-tuning evaluation on IN1K (see~\cref{tab:imagenetcompare}) shows that adopting MIM pre-training as the initialization improves the performance more significantly than CL pre-training. Thus, we hypothesize that \emph{lower layers matter much more than higher ones for the self-supervised pre-training methods to perform well on the data-sufficient IN1K}. In order to verify the hypothesis, we design another experiment by sweeping over the number of reserved consecutive leading blocks of pre-trained models for initialization (the remaining blocks after the reserved blocks are randomly initialized), and then fine-tuning the initialized models on IN1K (for the sake of simplicity, we adopt 100-epoch fine-tuning). The left figure of~\cref{fig:drop-transfer} shows that reserving only a certain number of leading blocks at the lower end can already achieve a significant performance gain over randomly initializing all the blocks (\ie, training from scratch) for both MAE and MoCo-v3 pre-training. Whereas, further reserving higher layers leads to marginal gains, which demonstrates our hypothesis.

\vspace{1.5mm}
\noindent\textbf{Analysis 2.} We further conduct similar experiments to the above setting on some small-scale downstream datasets, and find that \emph{higher layers matter in data-insufficient downstream tasks}. The results are shown in the remaining figures of~\cref{fig:drop-transfer}. We observe consistent performance improvements as the number of reserved pre-trained models' leading blocks increases. And the smaller the dataset scale, the more the performance benefits from the higher layers. It demonstrates that higher layers are still valuable and matter in data-insufficient downstream tasks. Furthermore, we observe comparable performance for the transfer evaluation of both MAE and MoCo-v3 pre-training when only a certain number of lower layers are reserved for initialization, while CL-based MoCo-v3 surpasses MIM-based MAE when higher layers are further reserved. It indicates that the higher layers of MoCo-v3 pre-trained models work better than MAE on data-insufficient downstream tasks, which is also consistent with our CKA-based analysis shown in~\cref{fig:cmp}, that CL pre-training learns more semantics at an abstract level relevant to recognition in higher layers (high similarity to the recognition-aligned reference model in higher layers). 

\subsection{Attention Map Analysis}
\label{Sec:attmap-analyses}

The attention maps can reveal the behaviors for aggregating information in the attention mechanism of ViTs, which are computed from the compatibility of queries and keys by dot-product operation. We introduce two metrics for further analysis based on the attention maps: (1) the attention distance, which is defined as (for the $j_{\mathrm{th}}$ token of the $h_{\mathrm{th}}$ head)
\begin{align}
    \mD_{h,j}=\sum_{i=1}^l \softmax(\mA_h)_{i,j}\mG_{i,j}~,
\end{align}
where $\mA_h\in\mathbb{R}^{l\times l}$ is the attention map for the $h_{\mathrm{th}}$ attention head, $\mG_{i,j}$ is the Euclidean distance between the spatial locations of the $i_{\mathrm{th}}$ and $j_{\mathrm{th}}$ tokens, and $l$ is the number of tokens; (2) the attention entropy, which is defined as
\begin{align}
    \mE_{h,j}=-\sum_{i=1}^l \softmax(\mA_h)_{i,j}\mathrm{log}(\softmax(\mA_h)_{i,j})~.
\end{align}
Then, the average attention distance and attention entropy across all tokens in the $h_{\mathrm{th}}$ head can be calculated as 
\begin{align}
    \bm{d}_{h}=\frac{1}{l}\sum_{j=1}^l \mD_{h,j}~~ \mathrm{and} ~~\bm{e}_{h}=\frac{1}{l}\sum_{j=1}^l \mE_{h,j}~.
\end{align}

\begin{figure*}[thbp!]
    \begin{minipage}[t]{1.0\textwidth}
    \centering
    \begin{tabular}{@{\extracolsep{\fill}}c@{}c@{}c@{}@{\extracolsep{\fill}}}
        \includegraphics[width=0.235\textwidth]{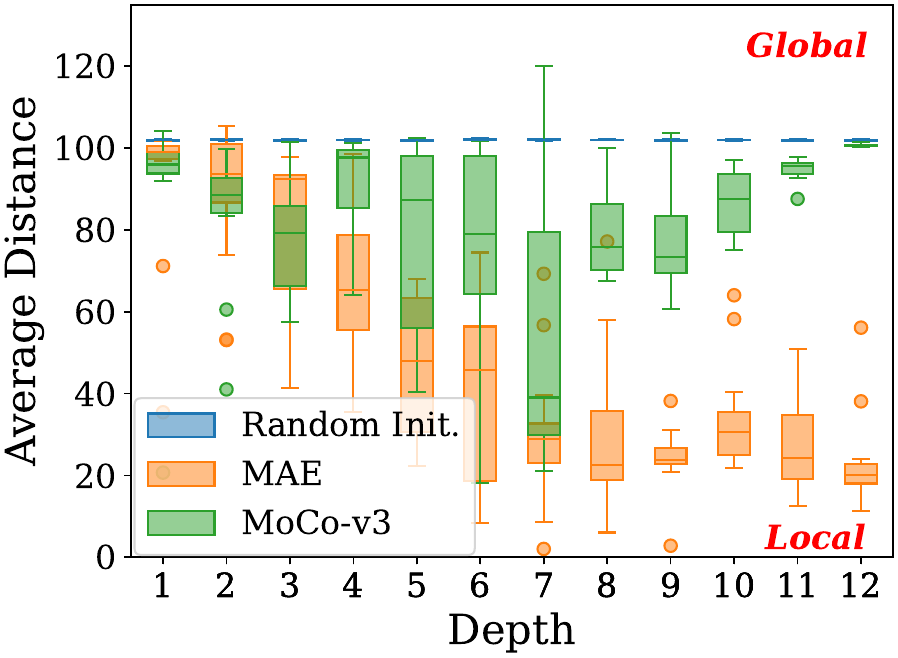}
        \hspace{0.003\textwidth}
        \includegraphics[width=0.235\textwidth]{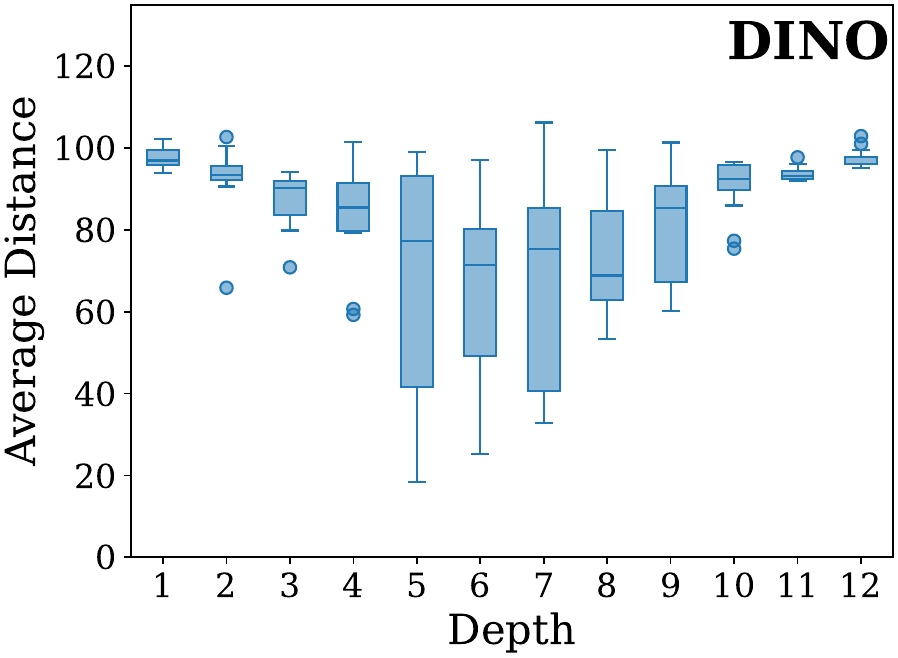}
        \hspace{0.003\textwidth}
        \includegraphics[width=0.235\textwidth]{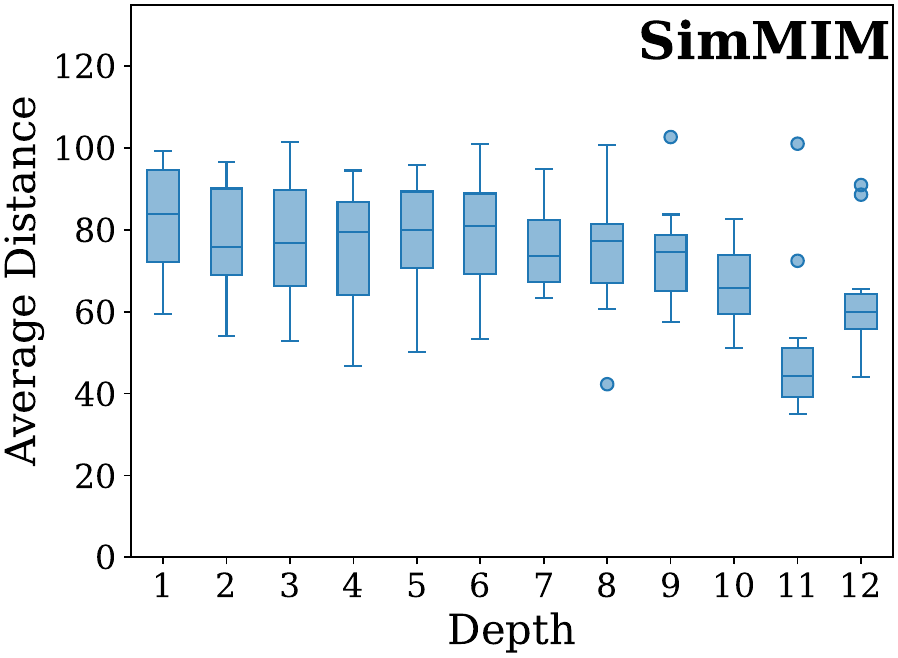}
        \hspace{0.003\textwidth}
        \includegraphics[width=0.235\textwidth]{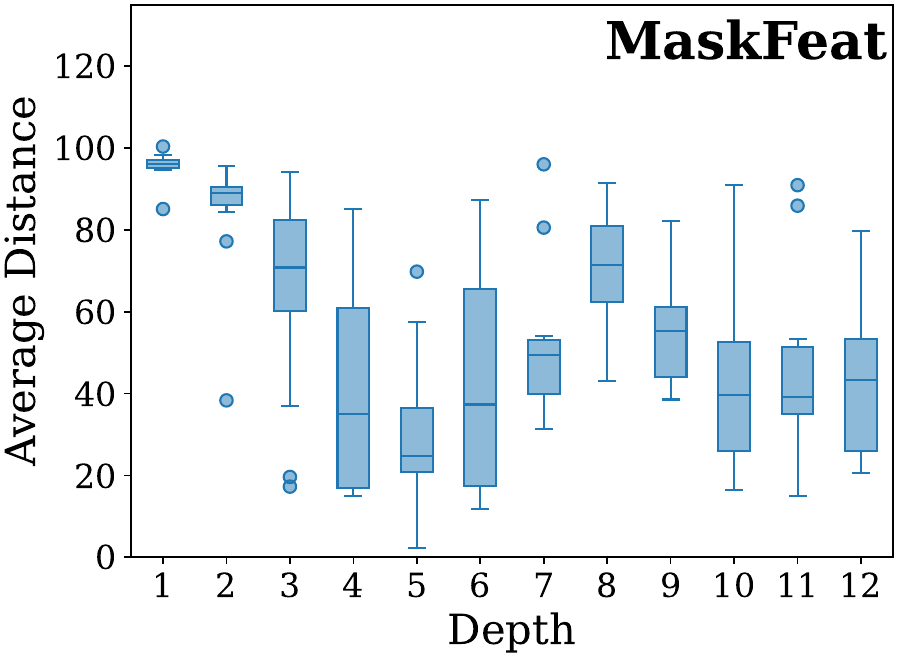} \\
        \includegraphics[width=0.235\textwidth]{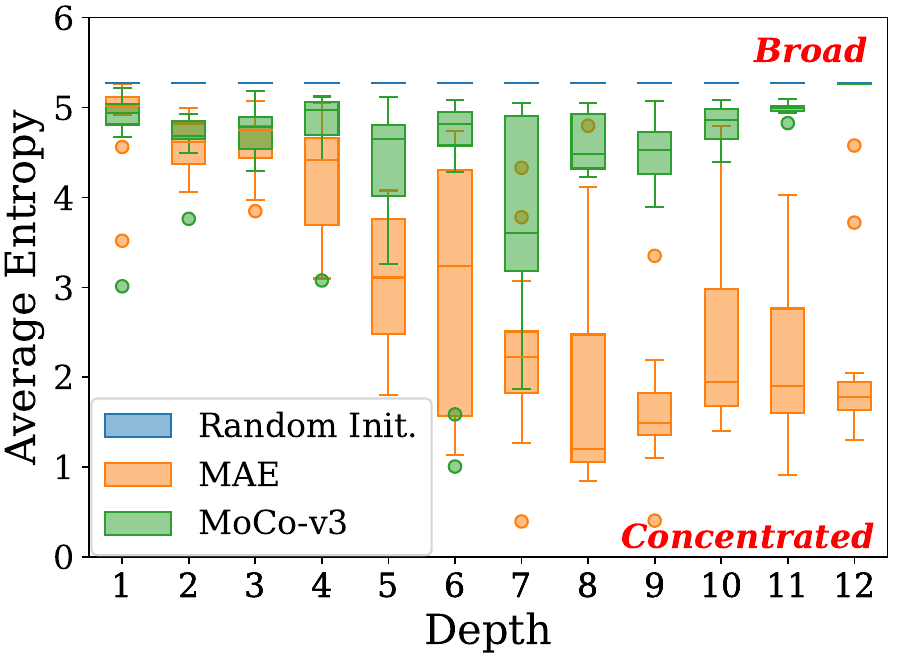}
        \hspace{0.003\textwidth}
        \includegraphics[width=0.235\textwidth]{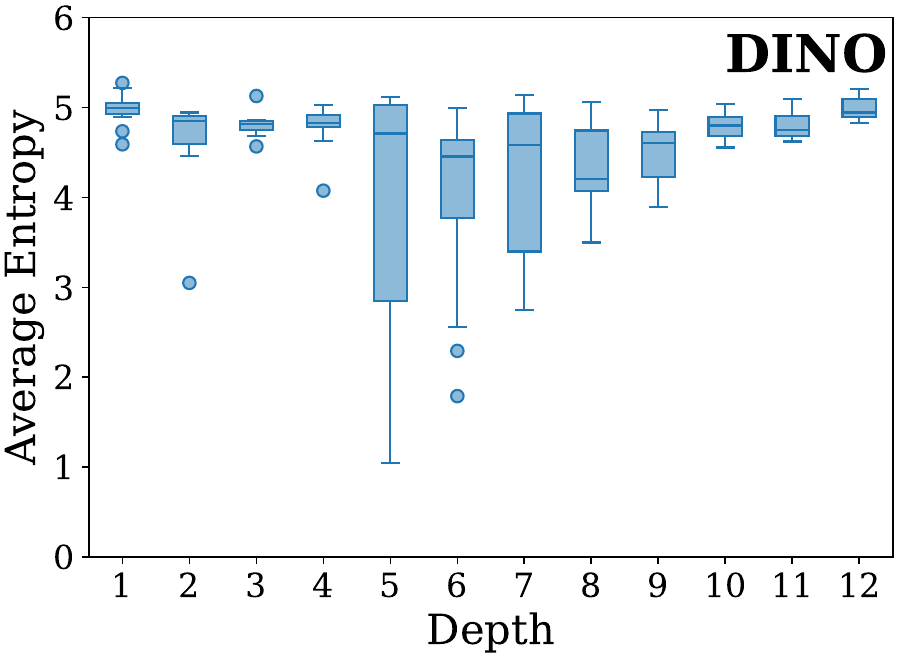}
        \hspace{0.003\textwidth}
        \includegraphics[width=0.235\textwidth]{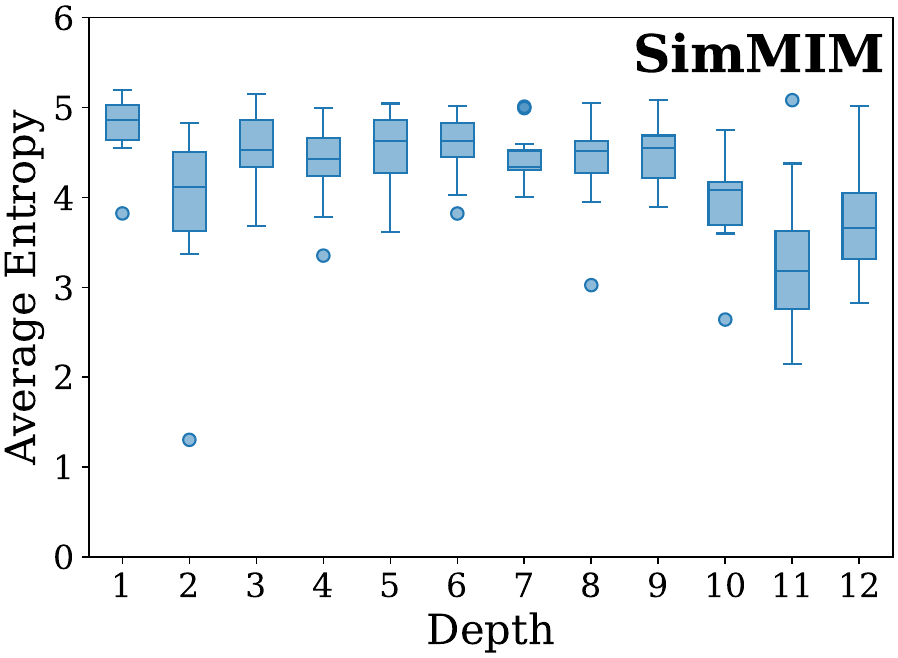}
        \hspace{0.003\textwidth}
        \includegraphics[width=0.235\textwidth]{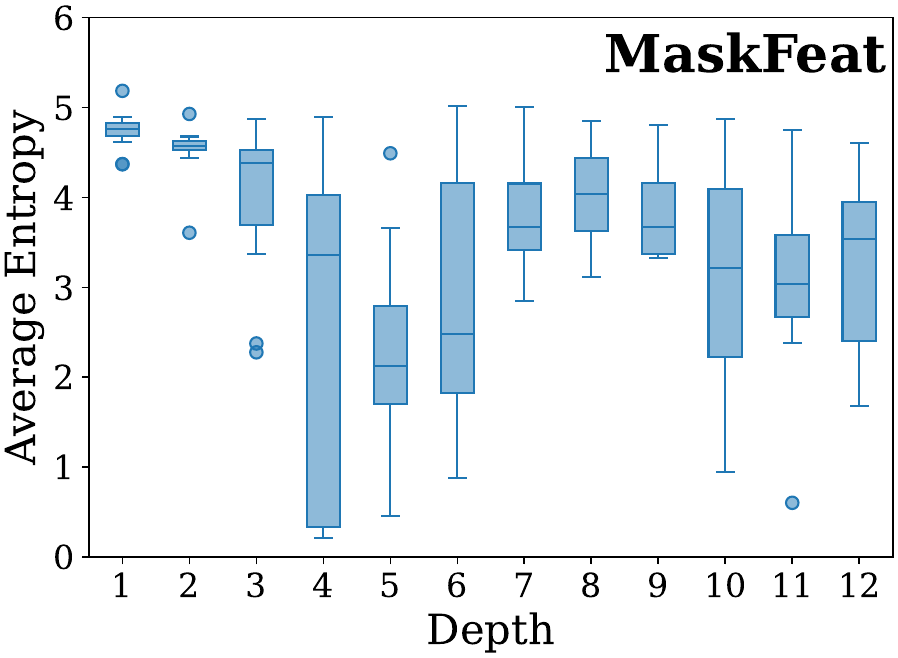}
    \end{tabular}
    \vspace{-0pt}
    \caption{\textbf{Attention distance and entropy analyses}. We visualize the layer-by-layer distributions of the average attention distance and entropy across all the different attention heads using the box-whisker plots. We specifically plot the MAE pre-trained and MoCo-v3 pre-trained ViT-Tiny models along with the randomly initialized one in the same figures (see left) for a more intuitive and compact comparison.
    }
    \label{fig:attn}
    \end{minipage}
    \vspace{-5pt}
\end{figure*}

Specifically, the average attention distance for one head reveals how much local \vs global information is aggregated, and a lower distance indicates that on average each token focuses more on neighbor tokens in this head. The average attention entropy for one head reveals the concentration of the attention distribution, and a lower entropy indicates that on average each token attends to fewer tokens. We use the models pre-trained with MoCo-v3, DINO, MAE, SimMIM and MaskFeat respectively for the attention map analysis in~\cref{fig:attn} and conduct the layer-by-layer analysis for the distributions of the average attention distance and entropy across all the different attention heads of the examined models.

\vspace{1.5mm}
\noindent\textbf{Analysis 3.} We first focus on the comparison between the MIM pre-trained and CL pre-trained models, trying to give some explanations for their diverse downstream performances observed in~\cref{sec:observation}. As shown in the left column of~\cref{fig:attn}, we observe that MoCo-v3 pre-training
generally results in more global and broad attention than MAE.
Even several leading blocks have a narrower range of attention distance and entropy than MAE. Compared with the randomly initialized model,
\emph{the MAE pre-trained model has significantly lower attention distance and entropy, which means that MAE pre-training noticeably introduces locality inductive bias}. This is consistent with the practice in~\cite{hiera} that the authors opt to teach the locality bias to the simple hierarchical ViT model Hiera using MAE pre-training while avoiding adding it through vision-specific modules like shifted windows or convolutions. As for the other pre-trained models in~\cref{fig:attn}, we can see that SimMIM also tends to make the model focus on local patterns with concentrated attention in higher layers like MAE, while DINO pre-training behaves like MoCo-v3 and results in broad and global attention in higher layers. It is noteworthy that the locality inductive bias does not mean that tokens in all attention heads attend to solely a few nearby tokens. The attention distance and entropy for different heads are still distributed in a wide range (except several last layers), which indicates that the heads have diverse specializations, making the models aggregate both local and global tokens with both concentrated and broad focuses. 

\begin{figure*}[thbp!]
    \begin{minipage}[t]{1.0\textwidth}
    \centering
    \begin{tabular}{@{\extracolsep{\fill}}c@{}c@{}c@{}@{\extracolsep{\fill}}}
        \includegraphics[width=0.235\textwidth]{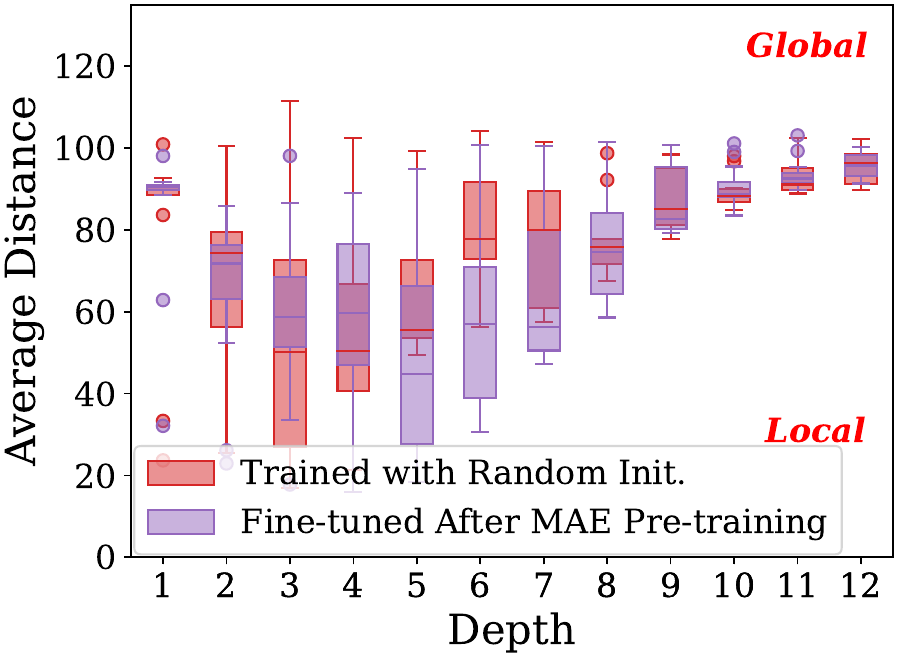}
        \hspace{0.003\textwidth}
        \includegraphics[width=0.235\textwidth]{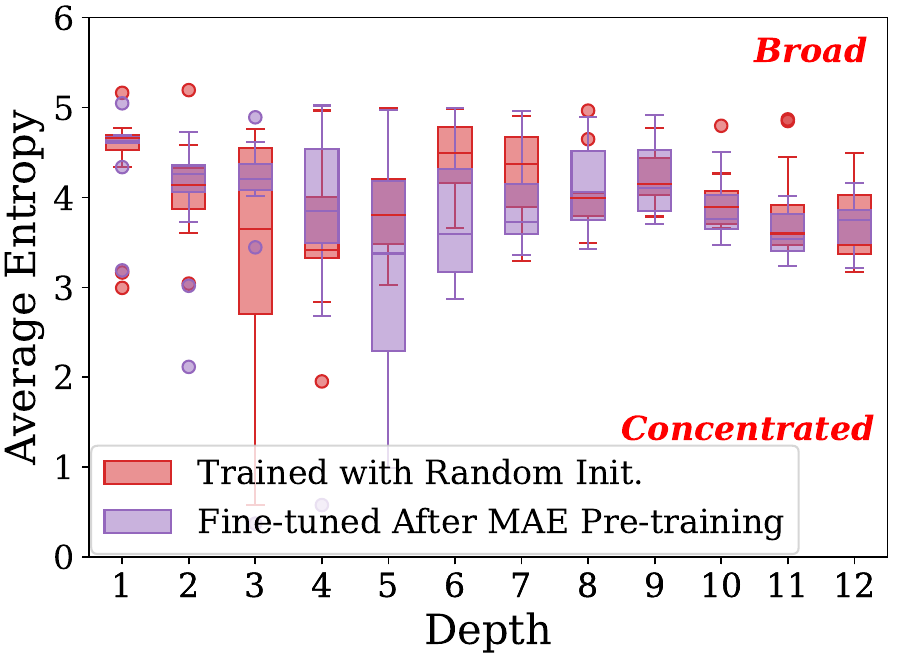}
        \hspace{0.003\textwidth}
        \includegraphics[width=0.235\textwidth]{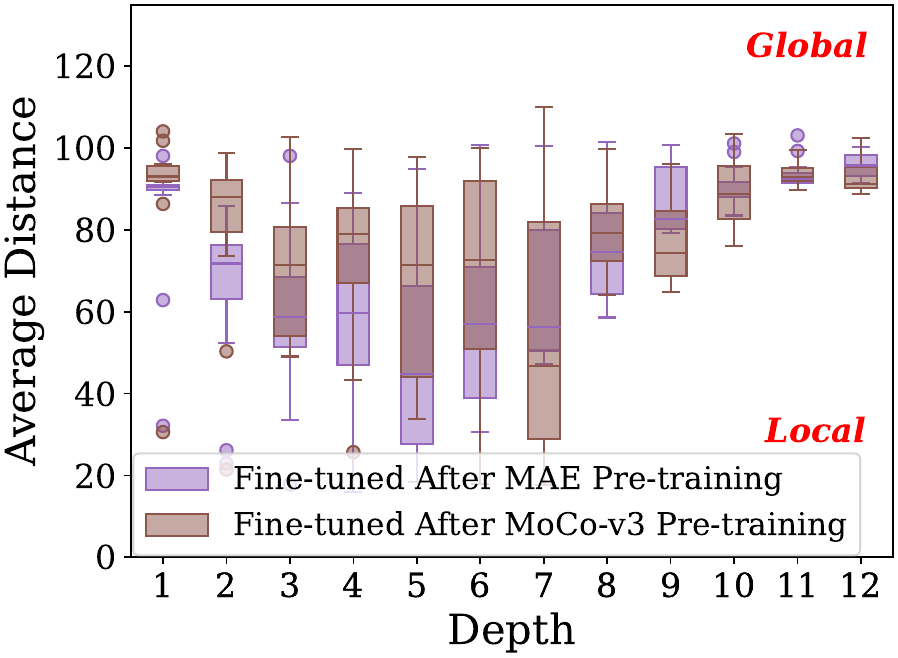}
        \hspace{0.003\textwidth}
        \includegraphics[width=0.235\textwidth]{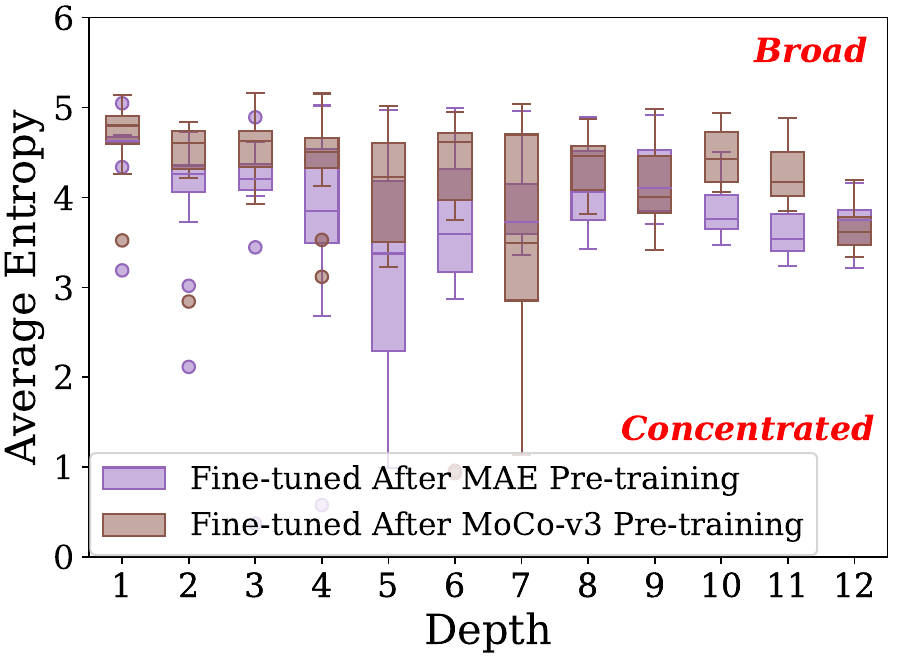}
    \end{tabular}
    \vspace{-0pt}
    \caption{\textbf{Attention distance and entropy analyses for the ViT-Tiny models fine-tuned on IN1K}. We compare the fine-tuned model after MAE pre-training with the fully-supervised one with random initialization in the left, while the comparison with MoCo-v3 is in the right.
    }
    \label{fig:attn_FT}
    \end{minipage}
    \vspace{-0pt}
 \end{figure*}

We then analyze how the entries in the left column of~\cref{fig:attn} behave when fine-tuned or trained on the downstream IN1K classification task. As shown in~\cref{fig:attn_FT}, we find that \emph{the pre-training with MAE makes the attention of its downstream fine-tuned model more local and concentrated than others.} 
First, we compare the fine-tuned model based on MAE pre-training with the fully-supervised one (random initialization) in the left of~\cref{fig:attn_FT}. We observe very similar attention behaviors between them, except that the attention with MAE pre-training 
is more local (with lower attention distance) and concentrated (with lower attention entropy) in middle layers compared with the fully-supervised one. 
Second, we conduct the attention comparison for the fine-tuned models with MAE and MoCo-v3 pre-training in the right of~\cref{fig:attn_FT}. It is shown that the fine-tuned model with MoCo-v3 pre-training still has more broad and global attention than the one with MAE pre-training in the middle layers.

As the MoCo-v3 pre-trained model generally has more global and broad attention than the MAE pre-trained one, we think this characteristic of MoCo-v3 makes its downstream fine-tuning take ``shortcuts", \ie, directly paying attention to global features and overlooking local patterns, which may be unfavorable for more fine-grained recognition when the downstream data for fine-tuning is sufficient. It leads to inferior downstream performance on IN1K.
As for the MAE pre-trained model, its distinct behaviors in higher layers with rather low attention distance and entropy may make it hard to be successfully transferred to data-insufficient downstream tasks, thus resulting in inferior performance on these tasks.



\section{Solution}
\label{Sec:solution}

\begin{figure*}[thbp!]
    \begin{minipage}[t]{1.0\textwidth}
    \centering
    \begin{tabular}{@{\extracolsep{\fill}}c@{}c@{}c@{}@{\extracolsep{\fill}}}
    \includegraphics[width=0.24\textwidth]{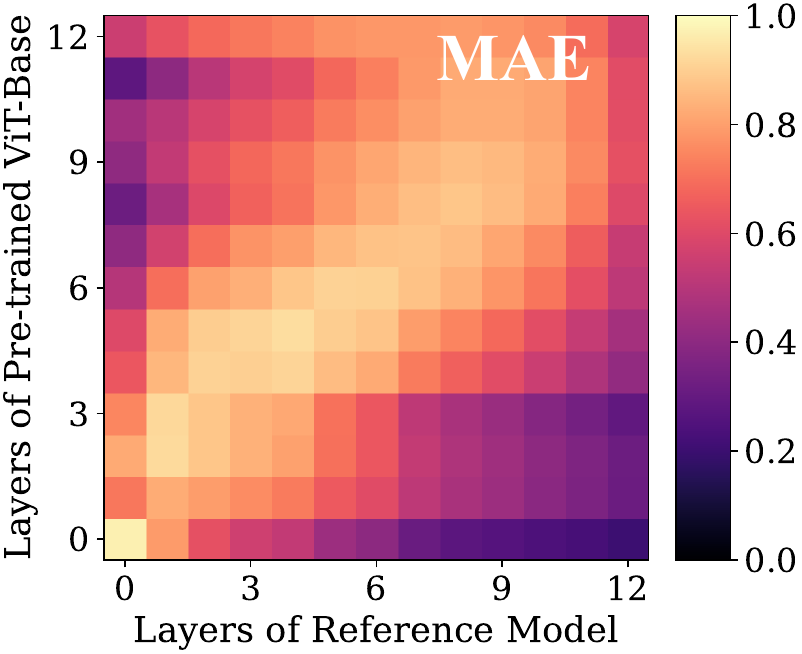}
    \hspace{0.006\textwidth}
    \includegraphics[width=0.26\textwidth]{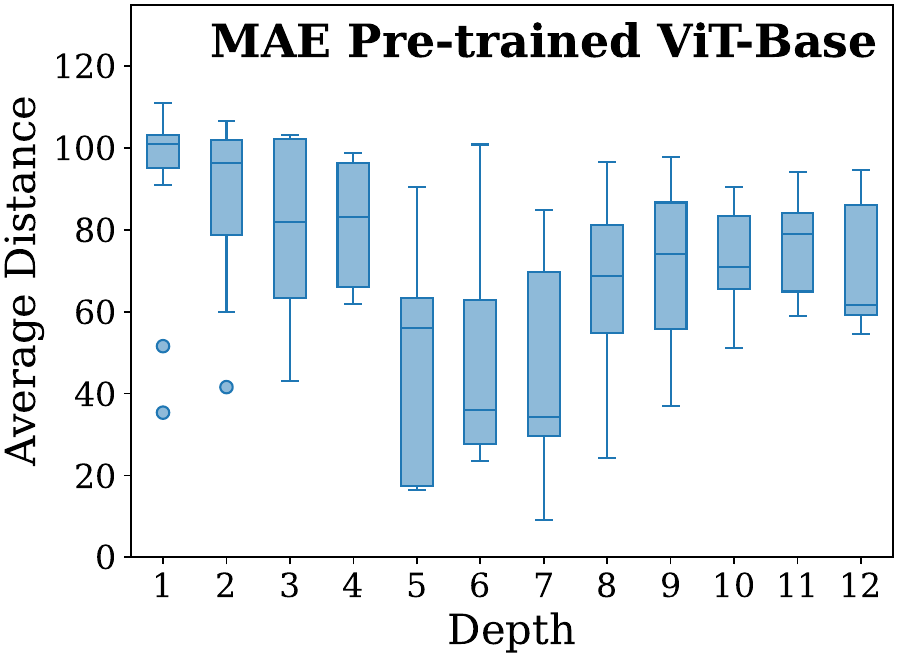}
    \hspace{0.006\textwidth}
    \includegraphics[width=0.26\textwidth]{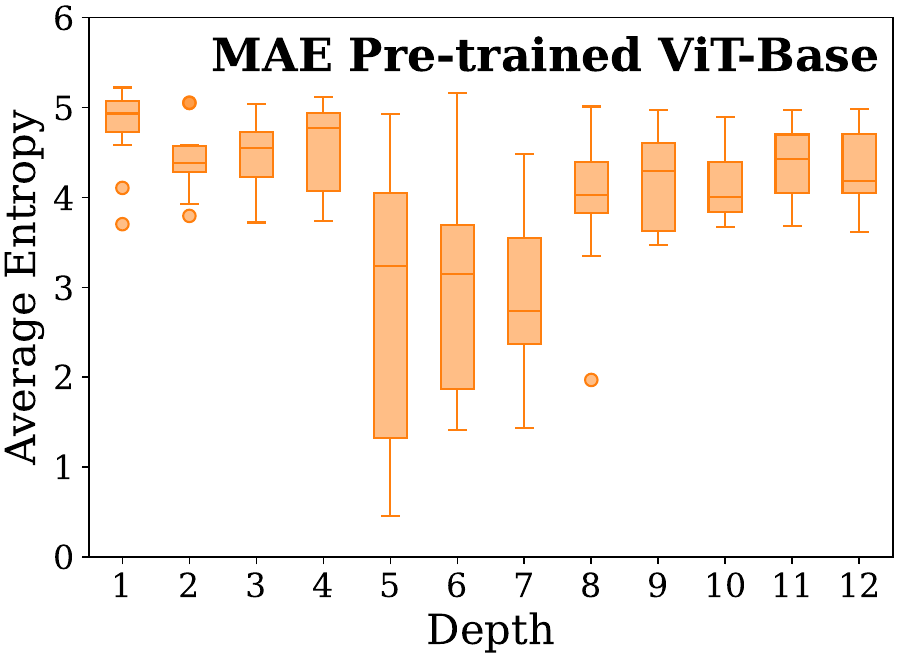}
    \end{tabular}
    \vspace{-0pt}
    \caption{The ViT-Base model with MAE pre-training~\cite{mae} can achieve a better alignment with our used reference model in~\cref{fig:cmp}. Its higher layers exhibit more global and broad attention than its lightweight ViT-Tiny counterpart with MAE pre-training.}
    \label{fig:mae-base}
    \end{minipage}
\vspace{-5pt}
\end{figure*}

\begin{figure*}[t]
    \begin{minipage}[t]{0.64\textwidth}
    \begin{tabular}{@{\extracolsep{\fill}}c@{}c@{}c@{}@{\extracolsep{\fill}}}
    \includegraphics[width=1.0\textwidth]{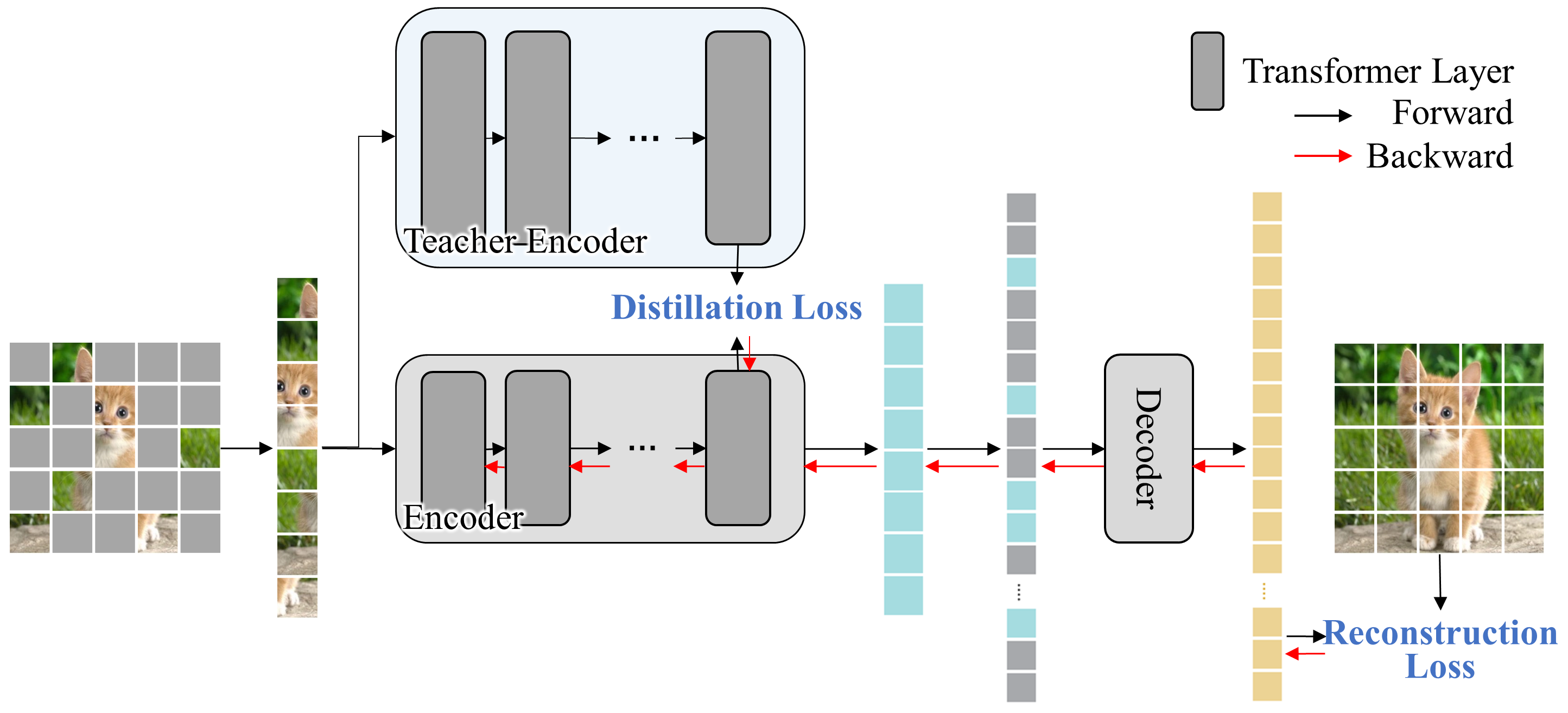}
    \end{tabular}
    \vspace{-0pt}
    \caption{\textbf{Illustration of our MAE pre-training with distillation on lightweight ViTs.} The teacher (MAE pre-trained ViT-Base) processes the same visible image patches as the student encoder.} 
    \label{fig:arch}
    \end{minipage} \hspace{2mm}
    \begin{minipage}[t]{0.33\textwidth}
    \begin{tabular}{@{\extracolsep{\fill}}c@{}c@{}c@{}@{\extracolsep{\fill}}}
    \includegraphics[width=1.0\textwidth]{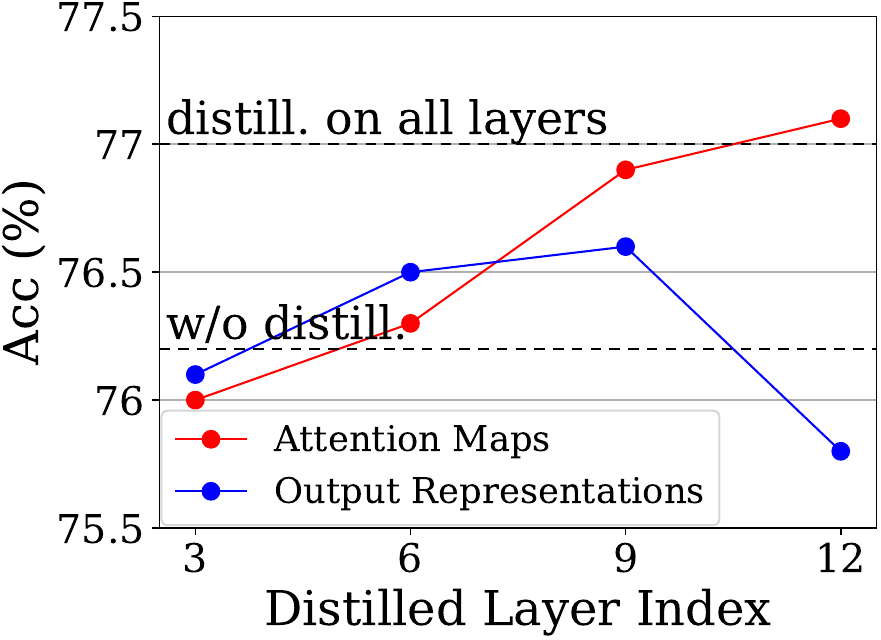}
    \end{tabular}
    \vspace{-0pt}
    \caption{Distillation on attention maps of higher layers significantly improves the performance over the pre-training baseline without distillation.}
    \label{fig:distill}
    \end{minipage}
\vspace{-0pt}
\end{figure*}

In the previous section, we have conjectured that it is hard for MIM pre-training, especially the MAE-based, to learn good representations relevant to recognition in higher layers, which results in unsatisfactory performance on data-insufficient downstream tasks, though providing good initialization only for lower layers contributes to notable performance improvement on downstream tasks with large-scale training data.
A natural question that arises is can the MIM pre-training gain more high-level semantic information by scaling up the models. We further examine a large ViT-Base model with MAE pre-training~\cite{mae} in~\cref{fig:mae-base}, and find it achieves a better alignment with our used reference model based on ViT-Tiny (the fully-supervised baseline based on our recipe in~\cref{fig:cmp}). It indicates that \emph{it is possible to extract features relevant to recognition in higher layers for the up-scaled encoder in MAE pre-training, and the pre-trained large model has more global and broad attention in higher layers.}
This observation motivates us to compress the knowledge of large pre-trained models to lightweight ones with knowledge distillation under the MIM framework (without relying on the training set labels), \eg, applying knowledge distillation during the MAE pre-training phase for lightweight ViTs. 

It is a common practice to utilize soft target probabilities (soft labels), features and intermediate representations to perform knowledge distillation. For instance, some works~\cite{distillbert, tinybert, minilmv2, mobilebert} perform distillation to obtain pre-trained compressed language models by exploiting soft target probabilities for masked language modeling predictions, embedding layer outputs, self-attention distributions or output representations (hidden states for the transformer blocks) from the teacher models to guide the student pre-training. As there are many distillation target alternatives and distillation strategies that could be explored, we only consider the most intuitive factors investigated in~\cref{Sec:Work} for distillation in our work, which are the layer-wise representations and attention maps. Our aim is to obtain good representations for the higher layers of lightweight ViTs to make them have high similarity to the recognition-aligned reference model (or the large pre-trained teacher model). We can thus distill the layer-wise representations explicitly or obtain the good representations implicitly through distilling the attention maps.


\subsection{Improve MAE Pre-Training Based on Distillation}
\label{Subsec:Distill}
We herein introduce our distillation process in the MAE pre-training framework (as illustrated in~\cref{fig:arch}) to showcase the effectiveness of our observation-analysis-solution flow in improving the MIM pre-traning of ViT-Tiny on higher layers.
We introduce our examined MAE pre-trained ViT-Base model in~\cref{fig:mae-base} as the teacher. During the MAE pre-training of ViT-Tiny, the teacher processes the same visible image patches as the student encoder ViT-Tiny, and two kinds of layer-wise attention-based and representation-based distillation losses can be constructed between the corresponding teacher's and student's layers for our investigation. The parameters of the student are updated based on the joint backward gradients from the distillation loss and the original MAE's reconstruction loss, while the teacher's parameters remain frozen throughout the pre-training process. The original MAE's reconstruction loss is constructed by using the output of the last layer of the student encoder as the decoder's input.

We construct the layer-wise attention-based and representation-based distillation losses as follows respectively based on the mean squared error ($\mathrm{MSE}$):
\begin{align}
    L_{\mathrm{attn}} = \mathrm{MSE}(\bm{A}_t, \bm{M}\bm{A}_s)~, \\
    L_{\mathrm{rep}} = \mathrm{MSE}(\bm{X}_t, \bm{X}_s\bm{N})~,
\end{align}
where $\bm{A}_t\in \mathbb{R}^{h\times l \times l}$ and $\bm{A}_s\in \mathbb{R}^{h'\times l \times l}$ refer to the attention maps of the corresponding teacher's and student's layers, with $h$ and $h'$ attention heads respectively. $l$ is the number of tokens. A learnable mapping matrix $\bm{M}\in \mathbb{R}^{h\times h'}$ is introduced to align the number of heads.
$\bm{X}_t\in \mathbb{R}^{l\times d}$ and $\bm{X}_s \in \mathbb{R}^{l\times d'}$ are the output representations of the corresponding teacher's and student's layers, with hidden dimensions as $d$ and $d'$ respectively. $\bm{N}\in \mathbb{R}^{d'\times d}$ is also a learnable mapping matrix. 

\begin{figure*}[thbp!]
    \begin{minipage}[t]{1.0\textwidth}
    \centering
    \begin{tabular}{@{\extracolsep{\fill}}c@{}c@{}c@{}@{\extracolsep{\fill}}}
    \includegraphics[width=0.24\textwidth]{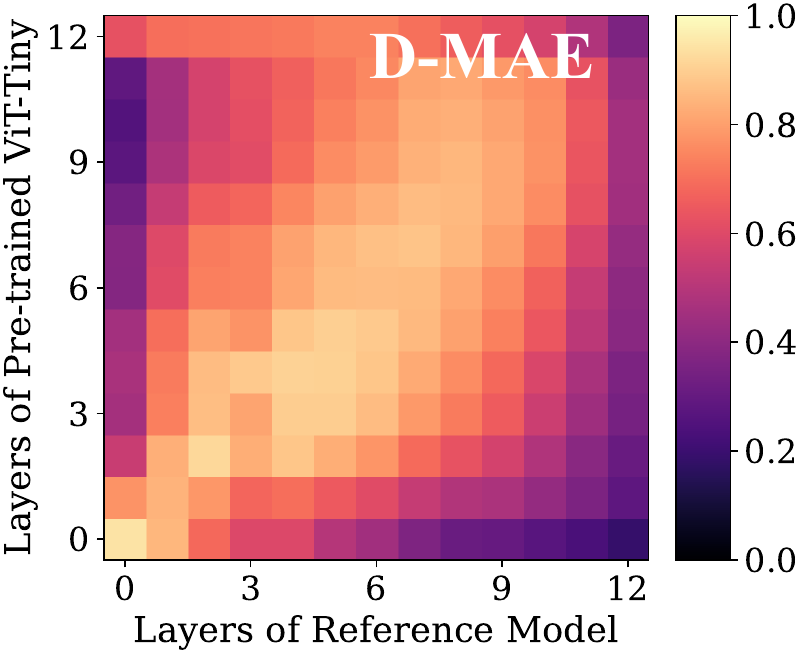}
    \hspace{0.008\textwidth}
    \includegraphics[width=0.26\textwidth]{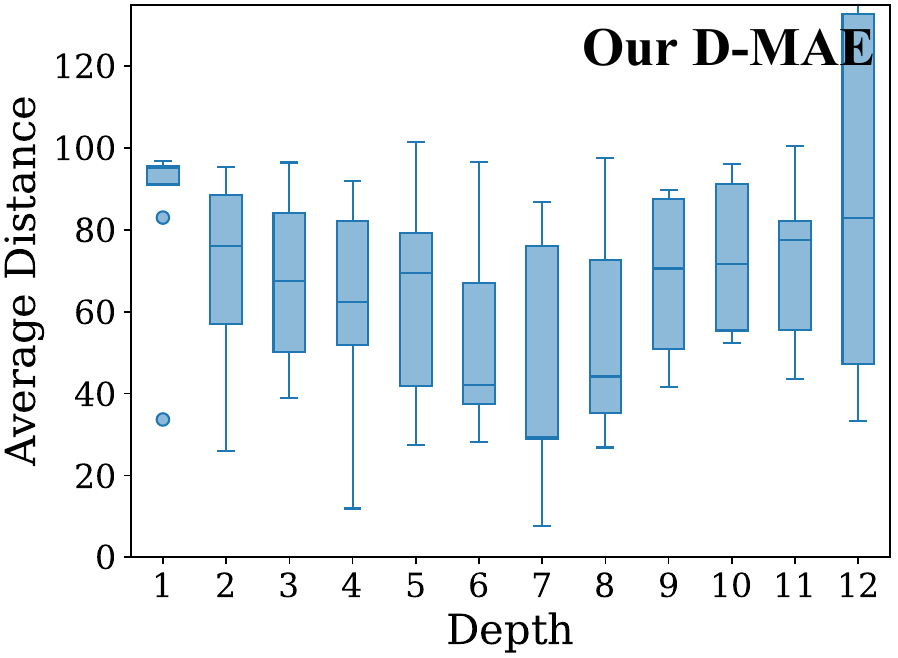}
    \hspace{0.008\textwidth}
    \includegraphics[width=0.26\textwidth]{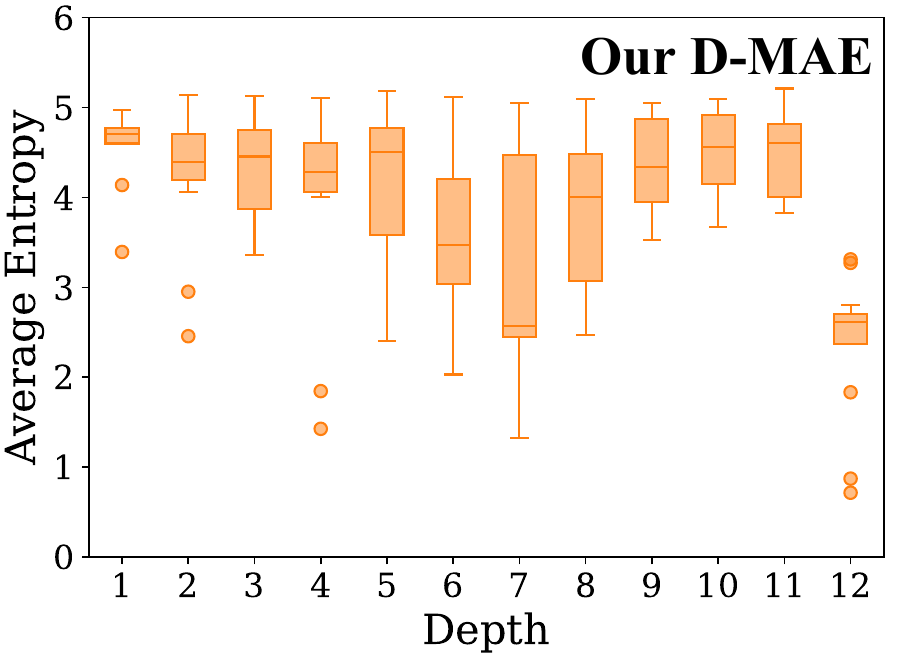} \vspace{-2pt}\\
    \includegraphics[width=0.24\textwidth]{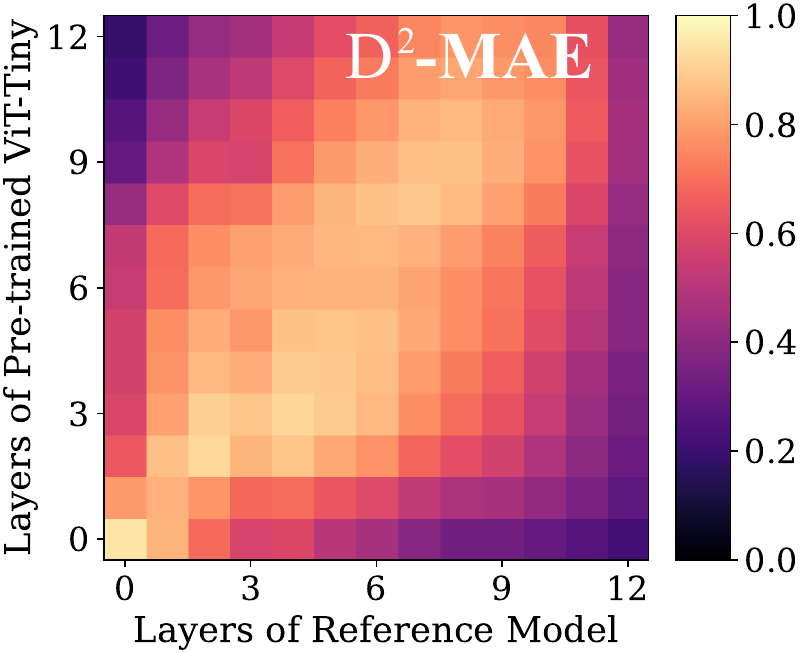}
    \hspace{0.008\textwidth}
    \includegraphics[width=0.26\textwidth]{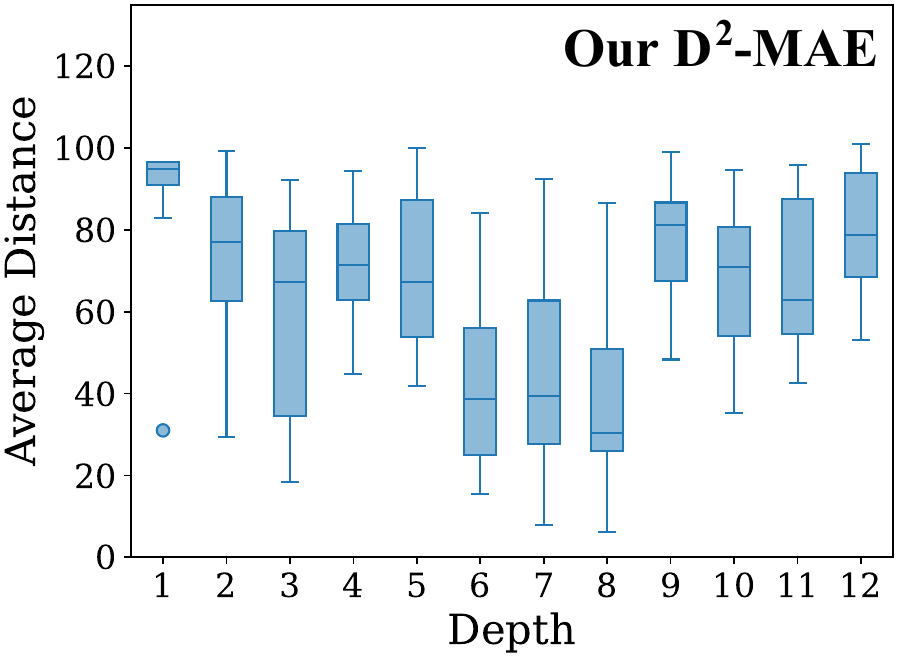}
    \hspace{0.008\textwidth}
    \includegraphics[width=0.26\textwidth]{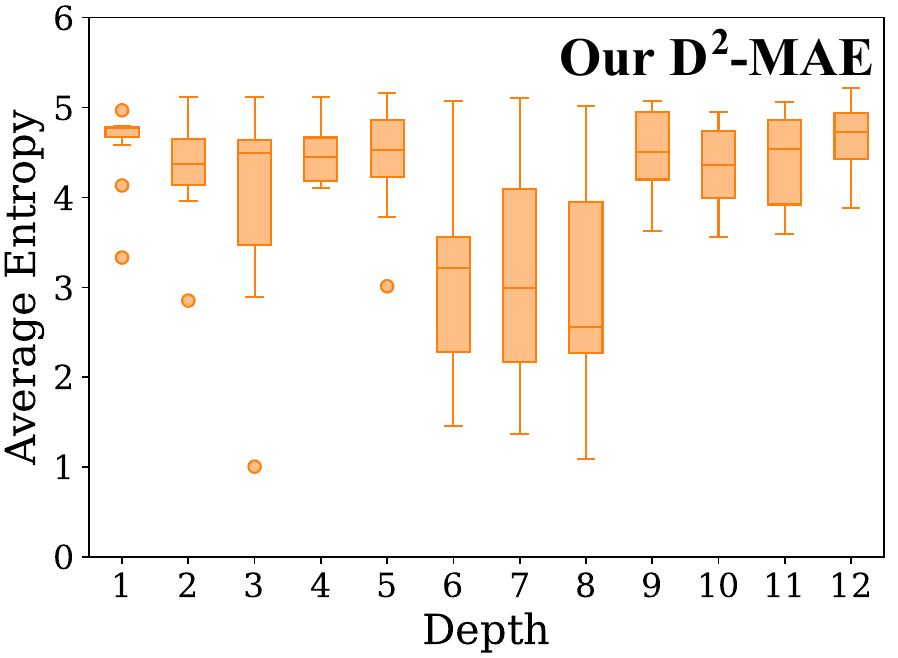}
    \end{tabular}
    \vspace{-0pt}
    \caption{\textbf{Distillation compresses the good representation of the ViT-Base teacher to the student through our D-MAE or D$^2$-MAE pre-training}. The distilled student shows higher layer-wise representation similarity to our used fully-supervised reference model in~\cref{fig:cmp}. Our D$^2$-MAE pre-training further improves the behavior patterns of the attention weights on higher layers, especially on the 12th layer.}
    \label{fig:cmp-distill}
    \end{minipage}
\vspace{-0pt}
\end{figure*}

\emph{Distill on attention maps or output representations?} We conduct a fine-tuning performance comparison (100 epochs for the sake of simplicity) on IN1K for the utilization of the two kinds of different distillation targets during our MAE pre-training (400 epochs). We test the distillation on the 3rd, 6th, 9th and 12th layers individually, \ie, the distillation losses are constructed between the pairs of layers at $1/4$, $2/4$, $3/4$ and $4/4$ depth of the teacher and student respectively. As shown in~\cref{fig:distill}, distillation on the attention maps of the last transformer blocks promotes the performance most, not only surpassing the same kind of distillation on lower layers, but also contributing to more performance gain in comparison to the distillation on the output representations. It is consistent with the analyses in~\cref{Sec:Work}. On the one hand, the lower layers learn good representations themselves during the MAE pre-training without distillation, and thus distillation on these layers contributes to marginal improvements, though distillation on the output representations directly performs a little better than on the attention maps of these layers. On the other hand, the higher layers rely on a good teacher to guide them to capture rich semantic features, and the attention weights on the higher layers of the teacher exhibit some patterns of behavior with much richer semantic knowledge, which may be the reason for the attention map's superiority over the representation for distillation on higher layers. \cref{fig:mae-base} also shows that the representations deteriorate on the 12th layer for the MAE pre-trained ViT-Base.
We denote the above best MAE pre-training based on attention map distillation on the last layer as D-MAE. As shown in the top-left of~\cref{fig:cmp-distill}, the CKA analysis for the D-MAE pre-trained ViT-Tiny indicates that the rich semantic knowledge of the teacher has been successfully transferred to the lightweight student.

\subsection{Decoupled Distillation Further Improves Pre-Training}
\label{Subsec:Decouple}

Despite the successfully obtained good representation with D-MAE, it is shown in~\cref{fig:cmp-distill} that the attention weights on the 12th layer of the obtained ViT-Tiny exhibit some weird behavior patterns, \ie, the distributions of the average attention distance and entropy cover a wide range of values while the entropy values are significantly lower than on other higher layers. That is to say the tokens in some heads of this last layer only attend to very few far away tokens while some focus on very few neighbor tokens. We think a reason for this is due to the conflict between distillation of high-level semantic knowledge and reconstruction of low-level pixels during MAE pre-training when these two tasks are both applied to the last layer, leading to a bad compromise for the different heads to learn the attention. To avoid this conflict, we propose a decoupled distillation strategy in this section.

\begin{table}[t]
\setlength{\tabcolsep}{3pt}
\centering
\caption{\textbf{Sweep over the intermediate layer index} to which the decoder for reconstruction is connected. Top-1 accuracy for classification tasks and AP for detection tasks are reported.
}
\label{tab:decouple}
\fontsize{8pt}{12pt}\selectfont
\begin{tabular}{cccccccc}
\toprule
\multirow{2}{*}{\emph{Datasets}} & \multicolumn{7}{c}{\emph{Encoder Layer Index for Reconstruction}} \\
& \textbf{w/o}$^{\bm{\diamond}}$ & \textbf{2} & \textbf{4} & \textbf{6} & \textbf{8} & \textbf{10} & \textbf{12} \\
 \midrule
\textbf{COCO}~(det.) & 39.6 & 41.8 & 42.2 & 42.2 & \textbf{42.5} & 42.1 & 42.0 \\
\textbf{Pets} & 84.5 & 87.6 & 89.5 & 89.9 & 90.4 & \textbf{91.2} & 89.1 \\
\textbf{Cars} & 80.3 & 86.0 & 87.6 & \textbf{88.4} & 88.2 & 87.9 & 87.5 \\
\textbf{CIFAR100} & 82.1 & 84.8 & 85.2 & \textbf{85.3} & 85.2 & 85.2 & 85.0 \\
\toprule
\end{tabular}
\begin{tablenotes}
    \footnotesize
    \item[$\bm{\diamond}$] indicates using distillation for pre-training without reconstruction.
\end{tablenotes}
\vspace{-0pt}
\end{table}

\emph{Decoupled distillation strategy.} As shown in ~\cref{fig:cmp}, the deterioration of representation learning on higher layers in the original MAE pre-training of lightweight ViTs indicates that there is some homogeneity among the representation learning of higher layers during the MAE reconstruction of the RGB pixel targets. That means it is not necessary to rely on the last layer for reconstructing the targets. We thus decouple the distillation task from the MAE reconstruction task by connecting the decoder for reconstruction to other intermediate layers of the student ViT-Tiny instead of the last layer during pre-training in~\cref{fig:arch}. We conduct a sweep over the intermediate layer index ranging from 2 to 12 for the optimal decoupled distillation. The fine-tuning results on some selected datasets (see~\cref{tab:decouple}) show that connecting to the 6th, 8th, or 10th layer can all further improve the performance over the model based on D-MAE (\ie, the column with the encoder layer index of 12 for reconstruction in~\cref{tab:decouple}). We use the 8th encoder layer for MAE reconstruction in the subsequent experimental analysis and validation by default, and denote this pre-training with \textbf{d}ecoupled \textbf{d}istillation as D$^2$-MAE. It can be seen in~\cref{fig:cmp-distill} that the attention weights on the higher layers after D$^2$-MAE pre-training exhibit very similar behavior patterns to the ViT-Base teacher (see~\cref{fig:mae-base}).

In~\cref{tab:decouple}, we also test the scenario of only relying on the attention map distillation to achieve good representation for the ViT-Tiny model, \ie, the MAE reconstruction loss in~\cref{fig:arch} is abandoned. It is shown that the experimental results without reconstruction are significantly lower than using both the distillation and reconstruction. However, only relying on the attention map distillation has already surpassed the MAE pre-training without distillation. This further demonstrates the effectiveness and superiority of our approach for developing the attention map distillation-based solution.

\begin{table*}[thbp!]
\setlength{\tabcolsep}{2.5pt}
\begin{center}
\renewcommand{\arraystretch}{0.9} 
{
\caption{\textbf{Distillation significantly improves performance on various tasks}. Top-1 accuracy is reported for classification tasks, mIoU is for ADE20K segmentation, AUC is for LaSOT tracking, and AP is for COCO detection and segmentation (Det./Seg.).
}
\label{tbl:transfer-distill}
\vspace{-0pt}
\fontsize{8pt}{12pt}\selectfont
\begin{tabular}{ccccccccccc}
\toprule
\multirow{2}{*}{\diagbox{\textbf{Methods}}{\textbf{Datasets}}} & \multicolumn{7}{c}{\textit{Classification Tasks}}&\multicolumn{3}{c}{\textit{Dense Prediction Tasks}} \\ \cmidrule(lr){2-8} \cmidrule(lr){9-11}
& \textbf{Flowers} & \textbf{Pets} & \textbf{Aircraft} & \textbf{Cars} & \textbf{CIFAR100} & \textbf{iNat18} & \textbf{IN1K} & \textbf{ADE20K} & \textbf{LaSOT} & \textbf{COCO}\\
\cmidrule(lr){1-8} \cmidrule(lr){9-11}
\colorgray{\textit{Supervised}~(IN1K)}\\
Our Recipe & 96.4 & 93.1 & 73.5 & 85.6 & 85.8 & 64.7 & 76.5 & 41.5 & 64.1 & 40.4/35.5\\
\cmidrule(lr){1-8} \cmidrule(lr){9-11}
\colorgray{\textit{MAE Self-Supervised}} \\
MAE & 85.8 & 76.5 & 64.6 & 78.8 & 78.9 & 64.0 & 78.0 & 34.5 & 61.6 & 39.9/35.4\\
D-MAE & 95.2 & 89.1 & 79.2 & 87.5 & 85.0 & 65.7 & 78.3 & 42.9 & 65.8 & 42.3/37.3\\
\rowcolor{lightlightlightgray}
D$^2$-MAE & 96.0 & 90.4 & 80.0 & 88.2 & 85.2 & 66.2 & 78.7 & 43.3 & 66.1 & 42.5/37.5\\
\rowcolor{lightlightgray}
$\Delta$ to MAE & \colorgreen{+10.2} & \colorgreen{+13.9} & \colorgreen{+15.4} & \colorgreen{+9.4} & \colorgreen{+6.3} & \colorgreen{+2.2} & \colorgreen{+0.7} & \colorgreen{+8.8} & \colorgreen{+4.5} & \colorgreen{+2.6/+2.1}\\
\rowcolor{lightlightgray}
$\Delta$ to Supervised & \colorred{-0.4} & \colorred{-2.7} & \colorgreen{+6.5} & \colorgreen{+2.6} & \colorred{-0.6} & \colorgreen{+1.5} & \colorgreen{+2.2} & \colorgreen{+1.8} & \colorgreen{+2.0} & \colorgreen{+2.1/+2.0}\\
\toprule
\end{tabular}}
\vspace{-0pt}
\end{center}

\vspace{-0pt}
\end{table*}

\subsection{Performance Improvements Based on Distillation}
\label{Subsec:distilledresults}

We finally evaluate our distillation-based MAE pre-trained models on all the classification and dense prediction tasks exploited in the observation section (\cref{sec:observation}). As shown in~\cref{tbl:transfer-distill}, our distillation contributes to significant performance improvements over the MAE pre-training without distillation, especially on the data-insufficient classification tasks and dense prediction tasks. In comparison to the supervised pre-training baseline (Our Recipe) relying on IN1K labels, our distillation-based MAE pre-training not only saves the costs for labeling the pre-training data, but also provides richer learning signals and avoids the bias towards the pre-defined set of IN1K labels. This leads to performance gains on Aircraft, Cars, iNat18, \etc., which may contain less overlapped categories or domains with ImageNet than Flowers, Pets and CIFAR100. The decoupled distillation strategy D$^2$-MAE also consistently improves over D-MAE on all the evaluated datasets. 

\section{State-of-the-art Comparison Results}
\label{Sec:EXP}
In this section, we mainly compare the obtained models, which are based on optimized pre-training strategies following our observation-analysis-solution flow, to the recent SOTA methods in the lightweight regime. We conduct the comparison on the ImageNet classification task, COCO object detection and instance segmentation tasks, ADE20K semantic segmentation task, and LaSOT single object tracking task. 

\begin{table*}[thbp!]

\begin{center}
    \setlength{\tabcolsep}{1.5pt}
    \caption{\textbf{Comparison results to the current SOTA lightweight models on ImageNet-1K.} Top-1 accuracy on its validation set with parameter count and latency are reported. All models are tested on input scale of 224$\times$224, except that MobileViT~\cite{mobilevit} and FastViT~\cite{fastvit} are tested with 256$\times$256 according to their original papers. `bs\emph{n}' means the latency is measured with batch Size \emph{n}.  
    \vspace{-0pt}
    }
    \fontsize{8pt}{12pt}\selectfont
	\begin{threeparttable}
	\begin{tabular}{llccccccccc}
    \toprule
    \multicolumn{1}{l}{\multirow{2}{*}{\textbf{Models}}} & \multicolumn{1}{l}{\multirow{2}{*}{\textbf{Pub.}}}  & \multicolumn{2}{c}{\emph{Pre-Training}} & \multicolumn{1}{c}{} & \multicolumn{2}{c}{\emph{Fine-Tuning}} & \multicolumn{1}{c}{\multirow{2}{*}{\textbf{Params}$\downarrow$}} & \multicolumn{2}{c}{\textbf{Latency}$\downarrow$} &  \textbf{ Top-1}$\uparrow$\cr
    \cmidrule(lr){3-4}\cmidrule(lr){6-7}\cmidrule(lr){9-10}
    & & \textbf{Methods} & \textbf{Data} & & \textbf{Distill.} & \textbf{Epochs} & & \emph{Orin(bs1)} & \emph{A100(bs16)} & (\%)\cr
    \cmidrule(lr){1-11}
    \colorgray{\textit{ConvNets Family}} \cr
    ResNet-18~\cite{resnet} & CVPR'16 & - & - & & \XSolidBrush & 100 & 11.7M & 0.99ms&1.12ms & 69.7 \cr
    EfficientNet-B0~\cite{efficientnet} & ICML'19 & - & - & & \XSolidBrush & 450 & 5.3M & 2.36ms&2.30ms & 77.7 \cr
    EfficientNet-B1~\cite{efficientnet} & ICML'19 & -  & - & & \XSolidBrush & 450 & 7.8M & 3.15ms&3.00ms & 78.8 \cr
    MobileNet-v3~\cite{mobilenetv3} & ICCV'19 & - & - & & \XSolidBrush & 600 & 5.5M & 1.55ms&1.63ms & 75.2 \cr
    MobileNet-v3$^{\bm{\dagger}}$~\cite{beyer2021knowledge} & CVPR'22 & Supervised & IN21K & & \Checkmark & 600 & 5.5M & 1.55ms&1.63ms & 77.9 \cr
    ConvNeXt V1-F~\cite{convnext} & CVPR'22 & - & - & & \XSolidBrush & 600 & 5.2M & 3.06ms&3.13ms & 77.5 \cr
    ConvNeXt V2-F$^{\bm{\dagger}}$~\cite{convnextv2} & CVPR'23 & FCMAE~\cite{convnextv2} & IN1K & & \XSolidBrush & 600 & 5.2M & 4.35ms&4.09ms & 78.5 \cr
    \cmidrule(lr){1-11}
    \colorgray{\textit{Non-Hierarchical ViTs}} \cr
    XCiT-T12/16~\cite{xcit} & NeurIPS'21 & - & - & & \Checkmark & 400 & 6.7M & 7.20ms&6.03ms & 78.6 \cr
    CaiT-XXS-24~\cite{cait} & ICCV'21 & -  & - & & \Checkmark & 400 & 12.0M & 8.96ms&7.25ms & 78.4 \cr
    DeiT-Tiny~\cite{deit} & ICML'21 & -  & - & & \XSolidBrush & 300 & 5.7M & 3.35ms&2.48ms & 72.2 \cr
    DeiT-Tiny$^{\bm{\dagger}}$~\cite{deit} & ICML'21 & -  & - & & \Checkmark & 1000 & 5.7M & 3.35ms&2.48ms & 76.6 \cr
    \cmidrule(lr){1-11}
    \colorgray{\textit{Hierarchical ViTs}} \cr
    LeViT-128~\cite{levit} & ICCV'21 & - & - & & \Checkmark & 1000 & 9.2M & 2.61ms&2.28ms & 78.6 \cr
    LeViT-192~\cite{levit} & ICCV'21 & -  & - & & \Checkmark & 1000 & 11.0M & 2.83ms&2.36ms & 80.0 \cr
    PiT-Ti~\cite{pit} & ICCV'21 & -  & - & & \Checkmark & 1000 & 5.1M & 3.80ms&2.85ms & 76.4 \cr
    Swin-1G~\cite{swin,mobileformer} & CVPR'22 & -  & - & & \XSolidBrush & 450 & 7.3M & 3.28ms&3.17ms & 77.3 \cr
    Mobile-Former-294M~\cite{mobileformer} & CVPR'22 & - & - & & \XSolidBrush & 450 & 11.4M & 9.71ms&6.86ms & 77.9 \cr
    EdgeViT-XS~\cite{edgevit} & ECCV'22 & - & - & & \XSolidBrush & 300 & 6.7M & 4.95ms&4.12ms & 77.5 \cr
    MobileViT-S~\cite{mobilevit} & ICLR'22 & -  & - & & \XSolidBrush & 300 & 5.6M & 3.93ms&3.72ms & 78.3 \cr
    FastViT-T12~\cite{fastvit} & ICCV'23 & - & - & & \XSolidBrush & 300 & 6.8M & 2.85ms&3.13ms & 79.1 \cr
    EfficientFormerV2-S1~\cite{efficientformerv2} & ICCV'23 & -  & - & & \Checkmark & 450 & 6.1M & 3.37ms&3.04ms & 79.7 \cr
    EfficientViT-M4~\cite{efficientvit-cvpr} & CVPR'23 & -  & - & & \Checkmark & 1000 & 8.8M & 4.06ms&2.90ms & 77.1 \cr
    SeaFormer-L~\cite{seaformer} & ICLR'23 & -  & - & & \XSolidBrush & 600 & 14.0M & 5.12ms&4.01ms & 79.9 \cr
 
    EfficientViT-B1~\cite{efficientvit-iccv} & ICCV'23 & -  & - & & \XSolidBrush & 300 & 9.1M & 2.42ms&2.40ms & 79.4 \cr
    \cmidrule(lr){1-11}
    \colorgray{\textit{Ours}} \cr
    \rowcolor{lightlightlightgray}
    & & - & - & & \XSolidBrush & 300 & 5.7M & 4.01ms & 3.20ms & 75.8 \cr
    \rowcolor{lightlightlightgray}
    &&MAE~\cite{mae}& IN1K & & \XSolidBrush & 300 & 5.7M & 4.01ms&3.20ms & 78.0 \cr
    \rowcolor{lightlightlightgray}
    &&D$^2$-MAE&  IN1K & & \XSolidBrush & 300 & 5.7M & 4.01ms&3.20ms & 78.7 \cr
    \rowcolor{lightlightlightgray}
    \multirow{-4}{*}{ViT-Tiny}& &D$^2$-MAE$^{\bm{\ddagger}}$& IN1K  & & \XSolidBrush & 1000 & 5.7M & 4.01ms&3.20ms & 79.4 \cr
    & & & & & & & & & & \vspace*{-10pt}\cr 
    \rowcolor{lightlightgray}
    & & - & - & & \XSolidBrush & 300 & 6.5M & 3.08ms&3.12ms & 75.8 \cr
    \rowcolor{lightlightgray}
    &&MAE~\cite{hiera}& IN1K & & \XSolidBrush & 300 & 6.5M & 3.08ms&3.12ms & 78.5 \cr
    \rowcolor{lightlightgray}
    \multirow{-3}{*}{Hiera-Tiny}&&D$^2$-MAE&  IN1K & & \XSolidBrush & 300 & 6.5M & 3.08ms&3.12ms & 78.9 \cr
    \bottomrule
	\end{tabular}
	\begin{tablenotes}
        \small
        \item[$\bm{\dagger}$] indicates that the corresponding models are further improved based on an additional pre-training stage for model initialization.
        \item[$\bm{\ddagger}$] indicates that our initialized ViT-Tiny additionally adopt the relative position embedding (RPE) trick used in~\cite{simmim} during fine-tuning. Note that Hiera-Tiny can benefit from neither this RPE trick nor the longer fine-tuning schedule.
    \end{tablenotes}
    \end{threeparttable}
    \label{tab:sota}
    \vspace{-0pt}
\end{center}

\vspace{-0pt}
\end{table*}

As we have stated in Observation 2 of~\cref{sec:observation}, our studying of MAE pre-training on lightweight ViTs is orthogonal to the network architecture design methodology in the ViT derivatives. We thus also apply our approach on the hierarchical vision transformer Hiera~\cite{hiera} to demonstrate the superiority of our observation-analysis-solution flow.
Hiera~\cite{hiera} is an extremely simple framework that is more accurate than previous models (\eg, the MViT family~\cite{mvitv1,mvitv2}) while being significantly faster both at inference and during training. Hiera avoids some redundant constraints used in MViTv2~\cite{mvitv2}, \ie, the convolutional structure and decoupled relative position encoding.
It also incorporates the local attention (\ie, mask unit attention) in the first two stages of the model and keeps the capability to perform sparse MAE pre-training (\ie, masked tokens are deleted). We resize the smallest model of Hiera in the original paper to a lightweight size to obtain the Hiera-Tiny with only a 6.5M parameter count.
Specifically, we only modify the initial embedding dimension to 46 to make it match the lightweight vanilla ViT model ViT-Tiny used in our paper, leaving other components including the number of heads, blocks, \etc unchanged.

Our observation and analysis on Hiera-Tiny are almost the same as that on ViT-Tiny, so we can directly apply our decoupled distillation on its MAE pre-training to improve the performance. It is noteworthy that we evaluate performance for Hiera-Tiny in this section only on the ImageNet classification task and COCO object detection and instance segmentation tasks following the original paper to achieve the image results due to its intrinsic hierarchical property. In Appendix~\ref{Appdix:appdix-distill}, we also show that our distillation technique D-MAE can help other-scale vanilla ViT students beyond ViT-Tiny to achieve better downstream performance.




\begin{figure*}[thbp!]
    \begin{minipage}{0.3\textwidth}
    \makeatletter\def\@captype{table} 
    \hvFloat[rotAngle=90,nonFloat=true,capWidth=w,capPos=top]%
{table}%
{\setlength{\tabcolsep}{13pt}
\fontsize{8pt}{12pt}\selectfont
    \begin{tabular}{lccccccccc}
        \toprule
        \textbf{Backbones} & \textbf{\#Param.} & \textbf{FLOPs} & \textbf{Pre-Train} & \textbf{AP$^{bb}$} & \textbf{AP$_{50}^{bb}$} & \textbf{AP$_{75}^{bb}$} & \textbf{AP$^{m}$} & \textbf{AP$_{50}^{m}$} & \textbf{AP$_{75}^{m}$} \\
        \midrule
        ResNet18~\cite{resnet} & 31M & 207G & Super. & 37.1 & 57.3 & 40.0 & 34.0 & 54.3 & 36.4 \\
        ResNet50~\cite{resnet} & 44M & 260G & Super. & 41.0 & 61.7 & 44.9 & 37.1 & 58.4 & 40.1 \\
        PVT-T~\cite{pvt} & 33M & 208G & Super. & 39.8 & 62.2 & 43.0 & 37.4 & 59.3 & 39.9 \\
        LightViT-T~\cite{lightvit} & 28M & 187G & Super. & 41.5 & \textbf{64.4} & 45.1 & 38.4 & \textbf{61.2} & 40.8 \\
        \rowcolor{lightlightlightgray}
        & & & Super. & 40.4 & 60.7 & 43.1 & 35.5 & 56.8 & 36.9 \\
        \rowcolor{lightlightlightgray}
        \rowcolor{lightlightlightgray}
        \multirow{-2}{*}{ViT-Tiny} & \multirow{-2}{*}{28M} & \multirow{-2}{*}{204G} & D$^2$-MAE & 42.5 & 62.6 & 45.6 & 37.5 & 59.4 & 39.4 \\
        \rowcolor{lightlightgray}
        & & & Super. & 37.9 & 58.7 & 40.4 & 34.3 & 55.5 & 35.9 \\
        \rowcolor{lightlightgray}
        \multirow{-2}{*}{Hiera-Tiny} & \multirow{-2}{*}{28M} & \multirow{-2}{*}{205G} & D$^2$-MAE & \textbf{43.2} & 64.0 & \textbf{46.7} & \textbf{38.7} & 60.7 & \textbf{41.4} \\
        \bottomrule
    \end{tabular}
}
{\textbf{Comparison results to previous SOTA lightweight models in the Mask R-CNN framework on COCO.}}%
{tab:cocosota}

    \end{minipage}
    \hspace{0.1\textwidth}
    \begin{minipage}{0.4\textwidth}
        \makeatletter\def\@captype{table}

\hvFloat[rotAngle=90,nonFloat=true,capWidth=w,capPos=top]%
{table}%
{\setlength{\tabcolsep}{1.9pt}
\vspace{-4pt}
\fontsize{8pt}{13.2pt}\selectfont
\begin{tabular}{lcccc}
    \toprule
    \textbf{Backbones} & \textbf{\#Param.} & \textbf{FLOPs} & \textbf{Pre-Train} & \textbf{mIoU} \\
    \midrule
    \multicolumn{3}{l}{\colorgray{\textit{High-resolution Segmentation-specific}}} \vspace*{3pt} \cr
    SeaFormer-L~\cite{seaformer} & 14.0M & 6.5G & Super. & 42.7 \\
    EfficientViT-B1~\cite{efficientvit-iccv} & 4.8M & 6.3G & Super. & 42.8 \\
    \midrule
    \multicolumn{5}{l}{\colorgray{\textit{Versatile Backbones with Semantic-FPN Decoder}}} \vspace*{3pt} \cr
    ResNet18~\cite{resnet} & 15.4M & 32.3G & Super. & 34.1 \\
    LeViT-192~\cite{levit} & 14.7M & 10.2G & Super. & 34.8 \\
    PVT-T~\cite{pvt} & 17M & 16.6G & Super. & 35.7 \\
    LightViT-T~\cite{lightvit} & 12.2M & 33.1G & Super. & 39.9 \\
    \rowcolor{lightlightlightgray}
    & & & Super. & 38.0 \\
    \rowcolor{lightlightlightgray}
    & & & D$^2$-MAE & 39.0 \\
    \rowcolor{lightlightlightgray}
    & & & Super.$^{\bm{\star}}$ &  41.5 \\
    \rowcolor{lightlightlightgray}
    \multirow{-4}{*}{ViT-Tiny} & \multirow{-4}{*}{7.0M} & \multirow{-4}{*}{13.5G} & D$^2$-MAE$^{\bm{\star}}$ & 42.8 \\
    \bottomrule
\end{tabular}
}%
{\textbf{Comparison results to lightweight semantic segmentation methods on ADE20K.}}%
{tab:ade20ksota}

        \vspace{0.08\textwidth}
        \makeatletter\def\@captype{table} 

\hvFloat[rotAngle=90,nonFloat=true,capWidth=w,capPos=top]%
{table}%
{\setlength{\tabcolsep}{1.8pt}
\vspace{-0pt}
\fontsize{8pt}{12pt}\selectfont
\begin{tabular}{llccccc}
    \toprule
    \textbf{Trackers} & \textbf{Backbones} & \textbf{\#Param.} & \textbf{FLOPs} & \textbf{Pre-Train} & \textbf{AUC} & \textbf{Speed}\\
    \midrule
    HiT-Base~\cite{hit} & LeViT-384 & 42.1M & 4.3G & Super. & 64.6 & 33\\
    HiT-Small~\cite{hit} & LeViT-128 & 11.0M & 1.1G & Super. & 60.5 & 72\\
    HiT-Tiny~\cite{hit} & LeViT-128S & 9.6M & 1.0G & Super. & 54.8 & 76\\
    LightTrack-L~\cite{lighttrack} & LT-Mobile  & 3.1M & 0.8G & Super. & 55.5 & - \\
    LightTrack-M~\cite{lighttrack} & LT-Mobile  & 2.0M & 0.5G & Super. & 53.8 & 41 \\
    E.T.Track~\cite{ettrack} & LT-Mobile & 7.0M & 1.7G & Super. & 59.1 & 47 \\
    FEAR-L~\cite{fear} & RegNet & 4.1M & 4.9G & Super. & 57.9 & - \\
    FEAR-XS~\cite{fear} & FBNet & 1.0M & 1.0G & Super. & 53.5 & 60 \\
    HCAT~\cite{hcat} & ResNet18 & 6.8M & 1.3G & Super. & 59.3 & 45 \\
    MixFormerV2-S~\cite{mixformerv2} & ViT-Base & 16.2M & 4.4G & Super. & 60.6 & 30 \\
    \rowcolor{lightlightlightgray}
    & & & & Super. & 64.1 & 41 \\
    \rowcolor{lightlightlightgray}
    & & & & D-MAE & 65.8 & 41 \\
    \rowcolor{lightlightlightgray}
    \multirow{-3}{*}{OSTrack-T} & \multirow{-3}{*}{ViT-Tiny} & \multirow{-3}{*}{8.0M} & \multirow{-3}{*}{1.9G} & D$^2$-MAE & 66.1 & 41 \\
    \bottomrule
\end{tabular}
}%
{\textbf{Comparison results to previous SOTA lightweight CPU-realtime trackers on LaSOT. We report CPU runtime speeds in FPS.}}%
{tab:lasotsota}

    \end{minipage}
    \hspace{0.05\textwidth}
    \begin{minipage}{0.1\textwidth}
    \makeatletter\def\@captype{table} 

\hvFloat[rotAngle=90,nonFloat=true,onlyText=true]
{table}
{
\begin{tabular}{c}
\multicolumn{1}{c}{\parbox{0.9\textheight}{\fontsize{8}{12}\selectfont $\bm{\star}$ indicates the models adopt an intermediate 1000-epoch fine-tuning on IN1K classification before ADE20K segmentation fine-tuning.}}\end{tabular}
}
{}{fig:resqrbd}
    \end{minipage}
 \end{figure*}

\vspace{1.5mm}
\noindent\textbf{Comparison on ImageNet Classification.}
In~\cref{tab:sota}, we compare both the enhanced ViT-Tiny and Hiera-Tiny with our distillation-based MAE pre-training to the current lightweight ConvNets and ViT derivatives. We report top-1 accuracy along with the model parameter count and the latency. By taking inspiration from the commonly used strategies in the network architecture design methodology, such as extending the supervised fine-tuning duration to a much longer schedule and using relative position bias, our approach can further benefit from them to achieve $79.4\%$/$78.9\%$ top-1 accuracy on ImageNet-1K, which is comparable to the current SOTA networks with sophisticated architecture design. It means that our proposed distillation strategy for the MAE pre-training can unleash the great potential of the \emph{extremely simple} lightweight ViTs, which is commonly overlooked in the past several years. Although some of the hierarchical ViTs (\eg, LeViT-192, EfficientFormerV2-S1 and EfficientViT-B1) enjoy higher accuracy and more attractive latency than ours simultaneously, our pre-training enables the more versatile ViT model ViT-Tiny to be outstanding on both the ADE20K segmentation and LaSOT tracking tasks in the lightweight regime (see below).

\vspace{1.5mm}
\noindent\textbf{Comparison on COCO Detection and Segmentation.}
For a more thorough study, we further conduct the SOTA comparison on the downstream object detection and segmentation tasks of COCO~\cite{coco}. As we have detailed in~\cref{Sec:Transfer}, we reproduce and rigorously follow the benchmarking setup in the Mask R-CNN framework in~\cite{li2021benchmarking} except for some minimal changes to the training settings. In specific, the different models in~\cref{tab:cocosota} are used for the initialization of the backbone. Our pre-training not only improves over the supervised pre-training significantly, but also makes the \emph{extremely simple} hierarchical Hiera-Tiny to outperform the delicately designed convolution-free ViT derivatives PVT-T and LightViT-T.

\vspace{1.5mm}
\noindent\textbf{Comparison on ADE20K Semantic Segmentation.}
We also conduct the SOTA comparison on the downstream semantic segmentation task of ADE20K~\cite{ade20k}. To align with the recent effort on reducing the computational cost of high-resolution dense prediction to ease its deploying on hardware devices, \eg, the segmentation-specific models SeaFormer and EfficientViT, we instead use the Semantic-FPN framework~\cite{semanticFPN} to evaluate the different lightweight versatile backbones in~\cref{tab:ade20ksota}. It shows that the image classification-proficient model LeViT with decreasing resolutions can not perform well on this ADE20K dense prediction task, while our pre-training with intermediate fine-tuning on IN1K enables ViT-Tiny to achieve comparable accuracy with the segmentation-specific models. 

\vspace{1.5mm}
\noindent\textbf{Comparison on LaSOT Visual Tracking.}
We finally conduct the SOTA comparison on the downstream single object visual tracking task of LaSOT~\cite{ade20k} in the lightweight regime, in which the real-time running on limited computing resources such as CPU is required. As we have detailed in~\cref{Sec:Transfer}, we reproduce and rigorously follow the original setup in the speed-oriented version of the SOTA tracker OSTrack~\cite{ostrack} (256$\times$256 input size of search patch) except that we only replace the backbone for joint feature extraction and relation modeling with different pre-trained ViT-Tiny, namely OSTrack-T in~\cref{tab:lasotsota}. It shows that our pre-training enhances OSTrack-T to outperform all the current SOTA lightweight CPU-realtime trackers and set a new record of $66.1\%$ AUC on LaSOT.

\section{Conclusion}

In this paper, we investigate an easily overlooked aspect of obtaining strong lightweight ViTs through MIM pre-training, which is optimizing the pre-training strategies for improving the \emph{extremely simple} lightweight ViTs and hence orthogonal to the delicate design of ViT derivatives by introducing sophisticated components with inductive biases. Our comprehensive analyses and investigations reveal several key factors that hide behind the achieved superior results, the value of which has also been demonstrated as it guides the design of our distillation strategy for better MIM pre-training. We believe our observations, analyses and solutions will improve the understanding of MIM pre-training for lightweight ViTs. We expect our observation-analysis-solution flow will further advance the SSL-based pre-training research.


\bibliography{mybib}
\clearpage

\begin{appendices}




\setcounter{equation}{0}
\setcounter{figure}{0}
\renewcommand \thefigure {S\arabic{figure}}
\renewcommand \theequation {S\arabic{equation}}
\setcounter{table}{0}
\renewcommand{\thetable}{S\arabic{table}}
%
%
%
%
\section{Experimental Details}\label{Appdix:appdix-detail}
\subsection{Recipe for Supervised Training of ViT-Tiny on IN1K}\label{Appdix:appdix-eval-detail}
We largely follow the common practice of supervised ViT training in~\cite{deit} except for some optimized hyper-parameters for augmentations (\ie, weaker regularization and augmentation than training large-scale models). The setting is detailed in~\cref{tbl:appdix-eval-detail}. We use the linear learning rate (\emph{lr}) scaling rule~\cite{imagenet_in_1_hour}: \emph{lr} = base \emph{lr}$~\times$~batch size / 256. We use layer-wise \emph{lr} decay following~\cite{beit, mae}.



\begin{table}[ht]
\begin{center}
\renewcommand{\arraystretch}{1.0} 
\setlength{\tabcolsep}{4pt}
\caption{Details of the recipe for supervised training of ViT-Tiny on IN1K.}
{
\small
\vspace{-0pt}
\begin{tabular}{cc}
\toprule
\textbf{Config.} & \textbf{Value}\\
\midrule
optimizer & AdamW\\
base \emph{lr} & 1e-3 \\
weight decay & 0.05 \\
optimizer momentum & $\beta_1, \beta_2=0.9, 0.999$ \\
batch size & 1024 \\
\emph{lr} schedule & cosine decay~\cite{sgdr} \\
warmup epochs & 5 \\
training epochs & 300 \\
augmentation & RandAug(10, 0.5)~\cite{randaug}  \\
colorjitter & 0.3 \\
label smoothing & 0 \\
mixup \cite{mixup} & 0.2 \\
cutmix \cite{cutmix} & 0 \\
drop path \cite{huang2016deep} & 0 \\
\bottomrule
\end{tabular}
\label{tbl:appdix-eval-detail}
}
\end{center}
\vspace{-0pt}
\end{table}


\subsection{Transfer Evaluation Details for Classification Tasks}\label{Appdix:appdix-transfer-detail}
We use 6 relatively smaller datasets than IN1K for transfer learning evaluation: Flowers~\cite{flower}, Pets~\cite{oxford-pet}, Aircraft~\cite{aircraft}, Cars~\cite{stanford-cars}, Cifar100~\cite{cifar}, and iNat18~\cite{inat}. For all these datasets except iNat18, we fine-tune with SGD (momentum=0.9), and the batch size is set to 512. The learning rates are swept over 3 candidates and the training epochs are swept over 2 candidates per dataset as detailed in~\cref{tbl:appdix-transfer-detail}. We adopt the cosine decay learning rate schedule~\cite{sgdr} with a linear warm-up. We resize images to 224 $\times$ 224. We adopt random resized crop and random horizontal flipping as augmentations and do not use any regularization (\eg, weight decay, dropout, or the stochastic depth regularization technique~\cite{huang2016deep}). For iNat18, we follow the same training configurations as those on IN1K.
\begin{table}[ht]
\setlength{\tabcolsep}{0.5pt}
\begin{center}
\renewcommand{\arraystretch}{1.0} 
{
\caption{Transfer evaluation details for some downstream classification tasks.}
\label{tbl:appdix-transfer-detail}
\vspace{-0pt}
\fontsize{8pt}{12pt}\selectfont
\begin{tabular}{cccc}
\toprule
\multirow{2}{*}{\textbf{Datasets}} & \multirow{2}{*}{\textbf{Learning rate}} & \textbf{\#Epochs} & \textbf{Layer-wise} \\
& & \textbf{(total, warm-up)} & \textbf{\emph{lr} decay}\\
\midrule
Flowers & \{0.01, 0.03, 0.1\} & \{(150,30),~(250,50)\} & \{1.0, 0.75\} \\
Pets & \{0.01, 0.03, 0.1\} & \{(70,14),~(150,30)\} & \{1.0, 0.75\}\\
Aircraft & \{0.01, 0.03, 0.1\} & \{(50,10),~(100,20)\} & \{1.0, 0.75\}\\
Cars & \{0.01, 0.03, 0.1\} & \{(50,10),~(100,20)\} & \{1.0, 0.75\}\\
CIFAR100 & \{0.03, 0.1, 0.3\} & \{(25, 5),~(50,10)\} & \{1.0, 0.75\}\\
\bottomrule

\end{tabular}}
\vspace{-0pt}
\end{center}
\end{table}

\section{More Analyses on Pre-Training}
\subsection{Analyses with More Models as Reference}\label{Appdix:appdix-ref}
\begin{figure*}[thbp!]
    \begin{minipage}[t]{1.0\textwidth}
    \centering
    \begin{tabular}{@{\extracolsep{\fill}}c@{}c@{}c@{}@{\extracolsep{\fill}}}
        \includegraphics[width=0.22\textwidth]{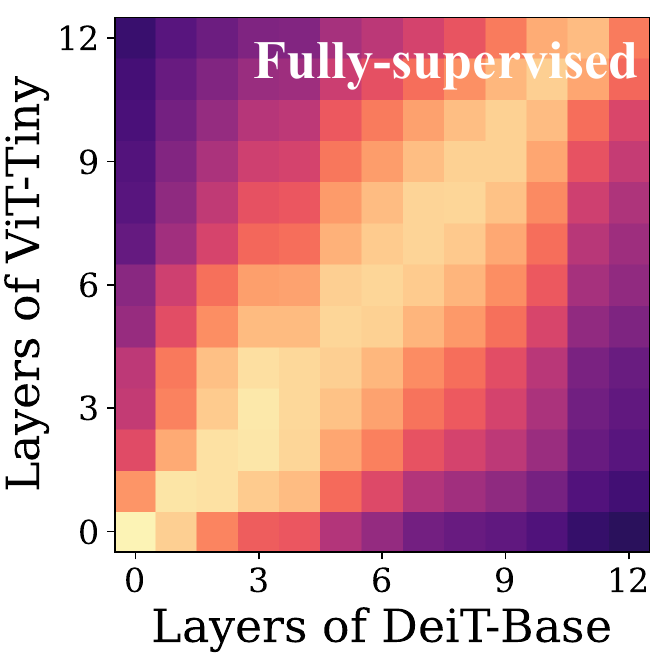}
        \hspace{0.005\textwidth}
        \includegraphics[width=0.22\textwidth]{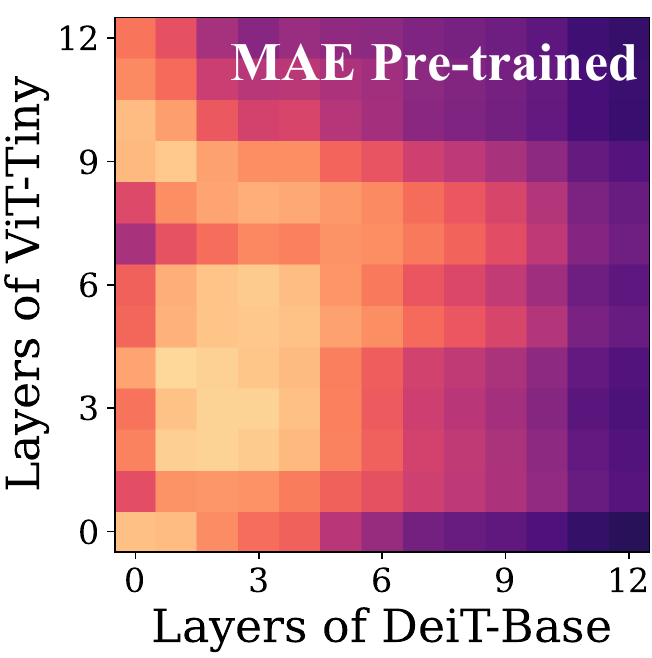}
        \hspace{0.005\textwidth}
        \includegraphics[width=0.22\textwidth]{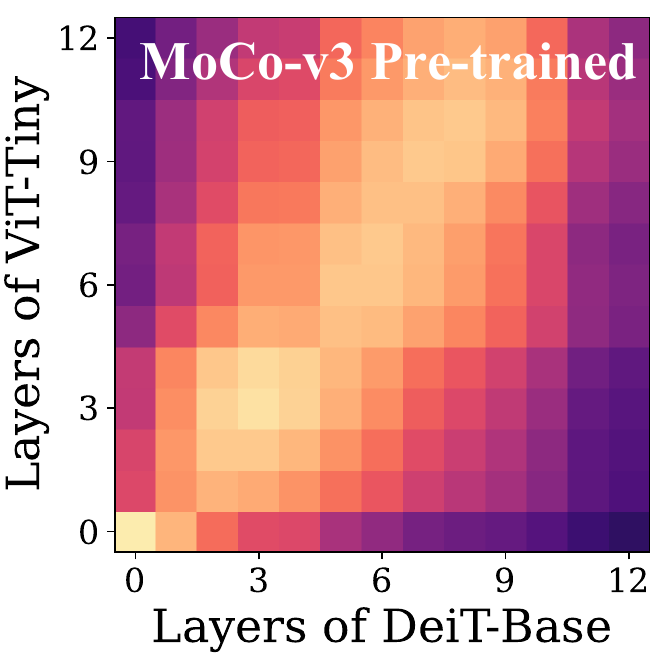}
        \hspace{0.005\textwidth}
        \includegraphics[width=0.27\textwidth]{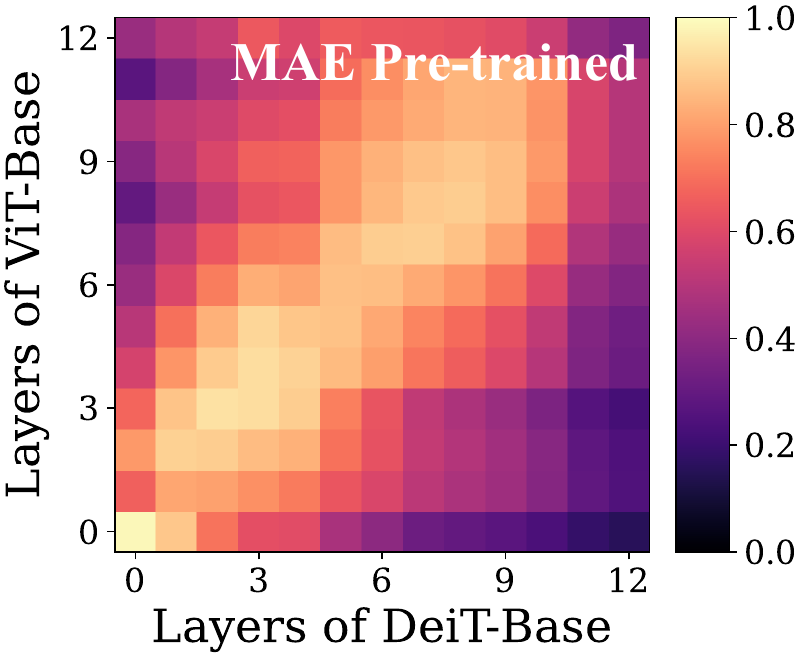} \\
        \includegraphics[width=0.22\textwidth]{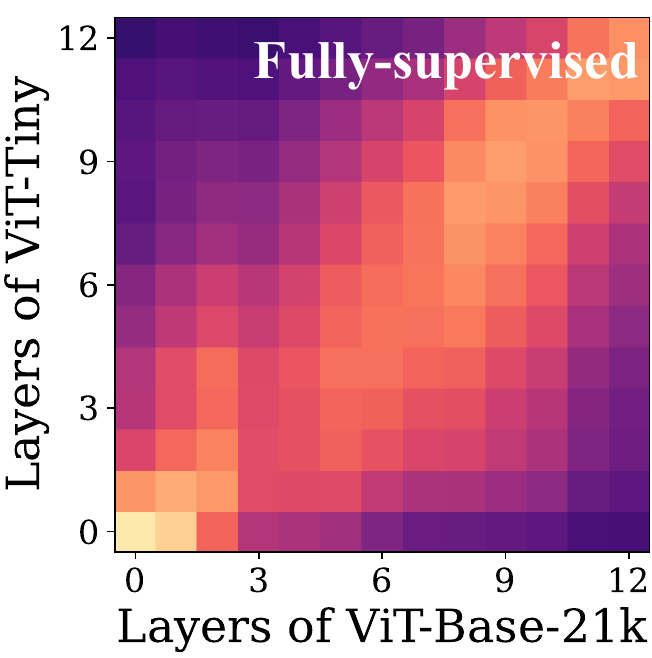}
        \hspace{0.005\textwidth}
        \includegraphics[width=0.22\textwidth]{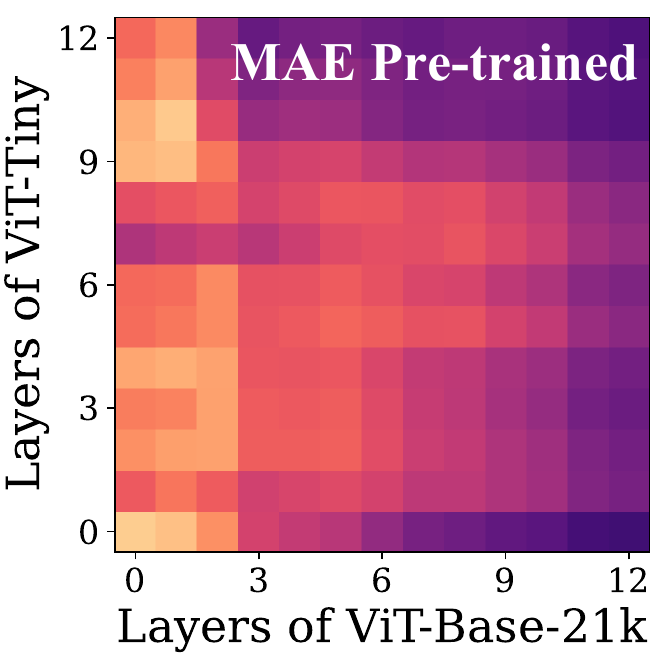}
        \hspace{0.005\textwidth}
        \includegraphics[width=0.22\textwidth]{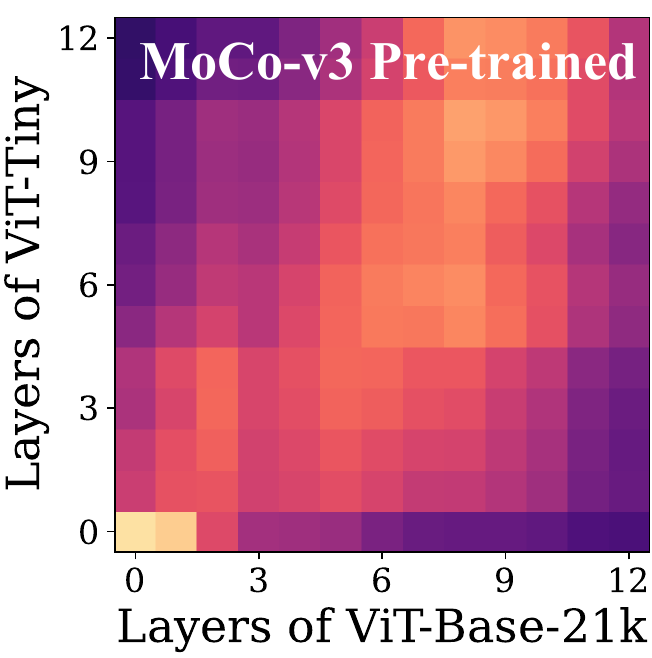}
        \hspace{0.005\textwidth}
        \includegraphics[width=0.27\textwidth]{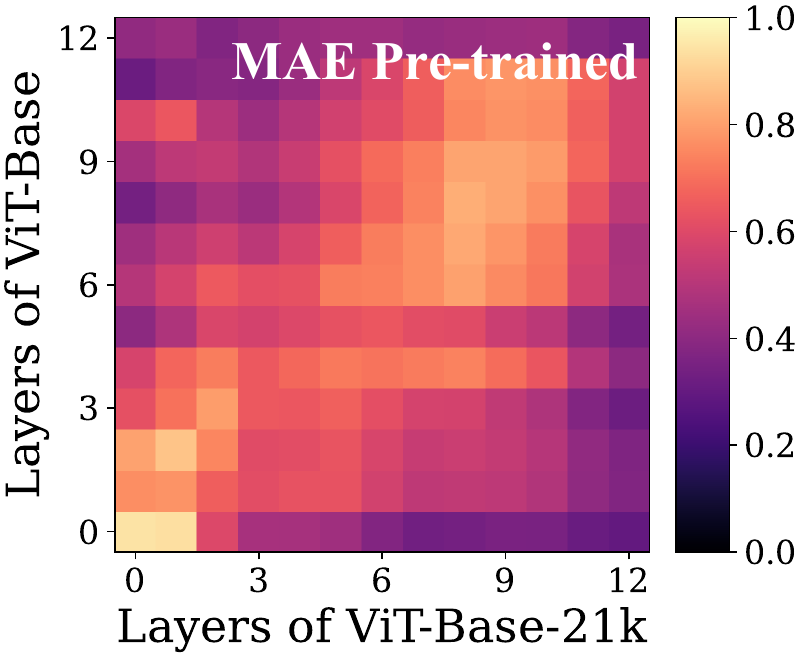} \\
        \includegraphics[width=0.23\textwidth]{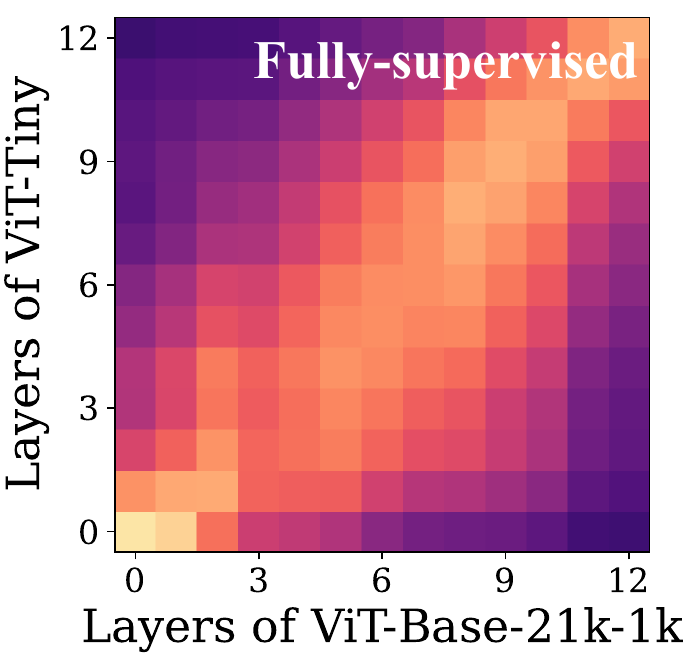}
        \includegraphics[width=0.23\textwidth]{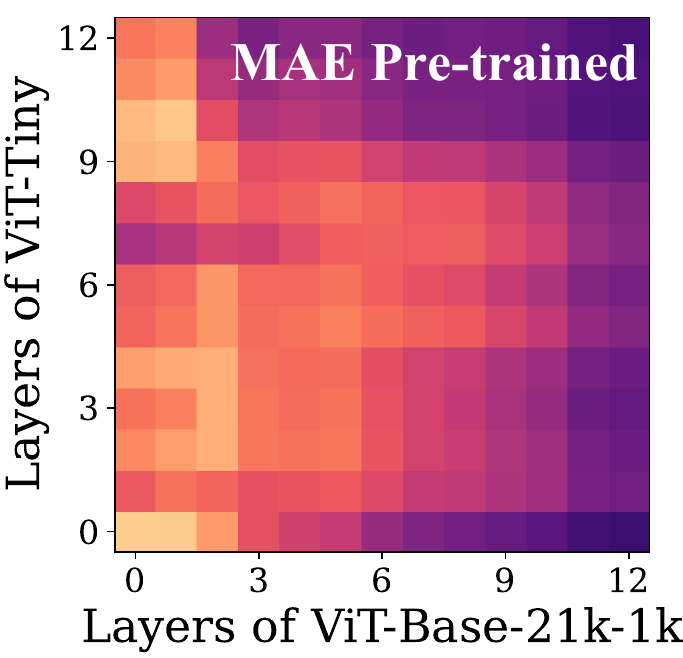}
        \includegraphics[width=0.23\textwidth]{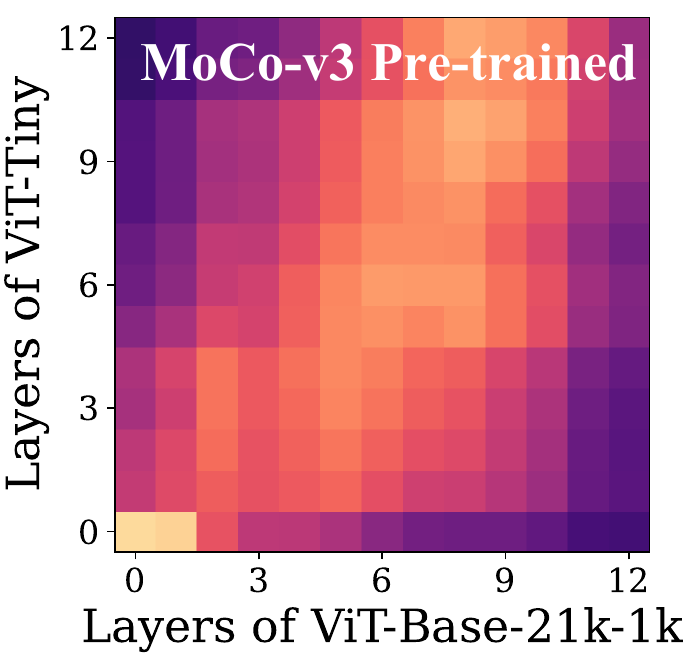}
        \includegraphics[width=0.27\textwidth]{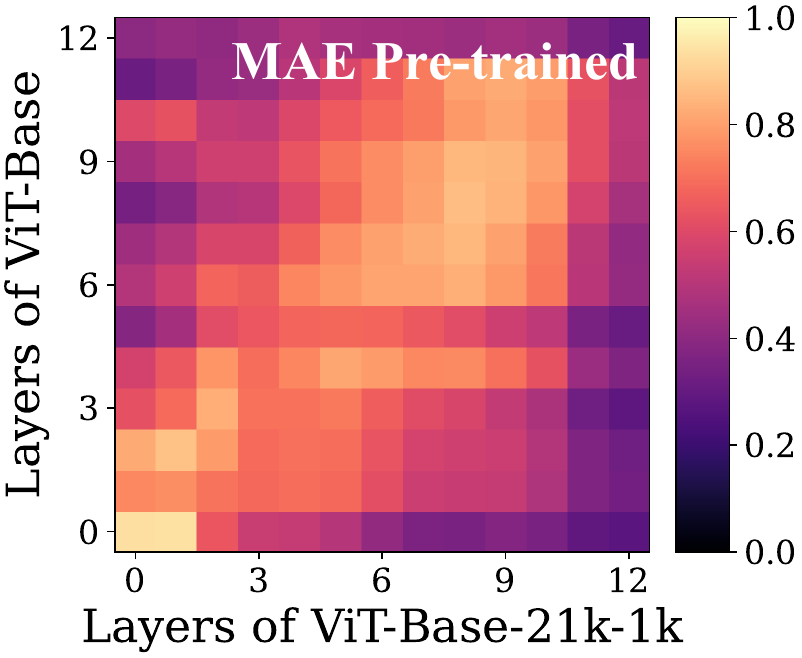} \\
    \end{tabular}
    \vspace{-0pt}
    \caption{\textbf{Layer representation analyses} with DeiT-Base (supervised training on IN1K following~\cite{deit} which achieves 82.0\% top-1 accuracy on IN1K), ViT-Base-21k (supervised pre-training on IN21K following~\cite{steiner2021train}), ViT-Base-21k-1k (first supervised pre-training on IN21K and then fine-tuning on IN1K following~\cite{steiner2021train}, achieving 84.5\% top-1 accuracy on IN1K) as the reference models.}
    \label{fig:appdix-cmp}
    \end{minipage}
    \vspace{0pt}
 \end{figure*}

In the main paper, the analyses based on the layer-wise representation similarity are mainly conducted by adopting the fully-supervised ViT-Tiny trained on the IN1K classification task based on our recipe (see~\cref{tab:imagenetcompare}) as the reference model. Here, we additionally introduce stronger recognition models as references to demonstrate the generalizability of our analyses. Specifically, we use ViT-Base models trained with various recipes as references, \eg, DeiT-Base (supervised training on IN1K following~\cite{deit} which achieves 82.0\% top-1 accuracy on IN1K), ViT-Base-21k (supervised pre-training on IN21K following~\cite{steiner2021train}), ViT-Base-21k-1k (first supervised pre-training on IN21K and then fine-tuning on IN1K following~\cite{steiner2021train}, achieving 84.5\% top-1 accuracy on IN1K). The layer representation similarity is presented in~\cref{fig:appdix-cmp}.

First, we observe that our used reference model in the main paper is aligned well with these larger models (as shown in the left column of~\cref{fig:appdix-cmp}). We conjecture that the supervised training generally makes the vanilla ViT models have similar layer representation structures. Based on these stronger reference models, we observe similar phenomena for MAE pre-training and MoCo-v3 pre-training of ViT-Tiny as discussed in the main paper, which demonstrates the robustness of our analyses and conclusions \wrt different reference models.

Then, we analyze the larger MAE pre-trained ViT-Base with these newly introduced models as references, as shown in the last column of~\cref{fig:appdix-cmp}. We observe that MAE pre-trained ViT-Base still aligns relatively well with these much stronger recognition models, which supports our claim in the main paper that \emph{it is possible to extract features relevant to recognition in higher layers for the up-scaled encoder in MAE pre-training}. It is the prerequisite for the improvement of the pre-trained ViT-Tiny based on the proposed distillation.

\section{More Analyses on Distillation}\label{Appdix:appdix-analyses}
\subsection{Applying Distillation on More Networks}\label{Appdix:appdix-distill}
To further evaluate our proposed distillation method, we additionally apply D-MAE to the pre-training of ViT-Small beyond ViT-Tiny using the MAE pre-trained ViT-Base as the teacher. The configurations of these models are presented in~\cref{tbl:appdix-vit}. The transfer evaluation results are presented in~\cref{tbl:appdix-small}. The transfer performance of the pre-trained ViT-Small based on D-MAE surpasses the baseline model without distillation during pre-training by a large margin, which shows the efficacy of the distillation following our observation-analysis-solution flow.
\begin{table*}[th]
\setlength{\tabcolsep}{6pt}
\begin{center}

\caption{
Configurations of vanilla ViTs.
}
\label{tbl:appdix-vit}
\vspace{-0pt}
\fontsize{8pt}{12pt}\selectfont
\begin{tabular}{lccccccc}
\toprule
\textbf{Models} & \textbf{Channels} & \textbf{\#Heads} & \textbf{\#Layers} & \textbf{\#Params} \\
\midrule
ViT-Tiny & 192 & 12 & 12 & 6M \\
ViT-Small & 384 & 12 & 12 & 22M \\
ViT-Base & 768 & 12 & 12 & 86M \\
\toprule
\end{tabular}
\item 
\end{center}
\end{table*}
\begin{table*}[th]
\setlength{\tabcolsep}{6pt}
\begin{center}

\caption{
Applying D-MAE to the pre-training of ViT-Small. Top-1 accuracy for the transfer performance on downstream classification tasks of pre-trained models w/ or w/o distillation is reported.
}
\label{tbl:appdix-small}
\vspace{-0pt}
\fontsize{8pt}{12pt}\selectfont
\begin{tabular}{cccccccc}
\toprule
\diagbox{\footnotesize\textbf{Methods}}{\footnotesize\textbf{Datasets}} & \textbf{Flowers} & \textbf{Pets} & \textbf{Aircraft} & \textbf{Cars} & \textbf{CIFAR100} & \textbf{iNat18} & \textbf{IN1K} \\
 \midrule
 \colorgray{\textit{Supervised (IN1K)}} \\
Our Recipe & \textbf{97.4} & \textbf{94.2} & 77.6 & 88.2 & \textbf{89.2} & 66.5 & 80.2 \\
 \midrule
\colorgray{\textit{Self-Supervised}} \\
MAE & 91.2 & 82.0 & 65.8 & 79.2 & 80.8 & 63.2 & 82.1 \\
\rowcolor{lightlightlightgray}
D-MAE & 95.8 & 91.4 & \textbf{80.7} & \textbf{88.3} & 87.8 & \textbf{66.9} & \textbf{82.5} \\
\rowcolor{lightlightgray}
$\Delta$ to MAE & \colorgreen{+4.6} & \colorgreen{+9.4} & \colorgreen{+14.9} & \colorgreen{+9.1} & \colorgreen{+7.0} & \colorgreen{+3.7} & \colorgreen{+0.4} \\
\toprule
\end{tabular}
\end{center}
\vspace{-0pt}
\end{table*}

\end{appendices}

\end{document}